\documentclass[10pt,twocolumn,letterpaper]{article}

\usepackage[pagenumbers]{cvpr}
\makeatletter
\patchcmd{\@maketitle}{\vskip .375in}{\vskip 6pt}{}{}
\patchcmd{\@maketitle}{\vspace*{12pt}}{\vspace*{0pt}}{}{}
\patchcmd{\@maketitle}{\vspace*{24pt}}{\vspace*{8pt}}{}{}
\makeatother
\usepackage{xcolor}
\usepackage{colortbl}

\usepackage{multirow}
\newcommand{\method}{\textsc{Lab2Sim}\xspace} 
\definecolor{plancolor}{RGB}{0,90,160}

\definecolor{cvprblue}{rgb}{0.21,0.49,0.74}
\usepackage[pagebackref,breaklinks,colorlinks,allcolors=cvprblue]{hyperref}

\hypersetup{
  pdftitle={Lab2Sim: Agentic Generation of Interactive Laboratory Equipment from Operation Videos},
  pdfauthor={Yuxin Lin, Pei Ren, Yirong Yang, Xuxin Lv, Xingchuang Liao, Diyuan Hou, Jiale He, Zhengping Che, Yongfei Zhang, Wenjun Wu, Zhaoxin Fan},
  pdfsubject={Research preprint}
}

\title{\method: Agentic Generation of Interactive Laboratory Equipment from Operation Videos}

\author{
  \begin{minipage}{0.96\textwidth}
    \centering
    \large
    Yuxin Lin$^{1}$ \quad Pei Ren$^{2,\dagger}$ \quad
    Yirong Yang$^{1}$ \quad Xuxin Lv$^{1}$ \quad
    Xingchuang Liao$^{1}$ \quad Diyuan Hou$^{1}$\\[2pt]
    Jiale He$^{1}$ \quad Zhengping Che$^{2}$ \quad
    Yongfei Zhang$^{1}$ \quad Wenjun Wu$^{1,*}$ \quad
    Zhaoxin Fan$^{1,*}$\\[5pt]
    {\small
      $^{1}$Beihang University, Beijing, China\\
      $^{2}$Beijing Humanoid Robot Innovation Center, Beijing, China\\[3pt]
    }
    {\footnotesize
      $^{*}$Corresponding authors \quad
      $^{\dagger}$Project leader \quad
      Emails: \texttt{\{zhaoxinf, wwj09315\}@buaa.edu.cn}
    }
  \end{minipage}
}
\date{}

\begin{document}
\twocolumn[{
\maketitle
\vspace{-2.2em}
\begin{center}
\includegraphics[width=\textwidth]{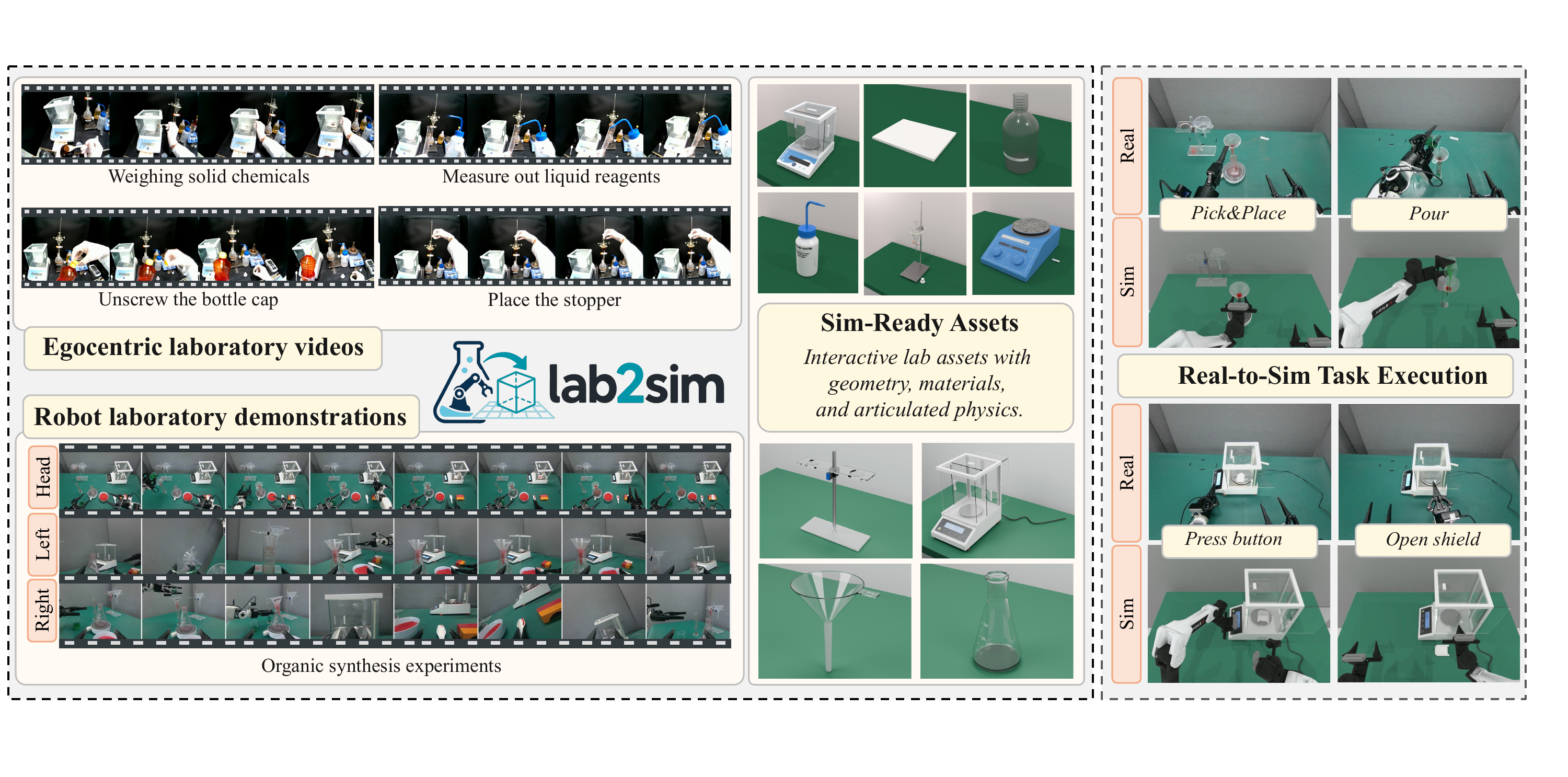}
\vspace{-0.3in}
\refstepcounter{figure}\label{fig:teaser}
\parbox{\textwidth}{\small\textbf{Figure~\thefigure.} \textbf{\method turns lab operation videos into simulation-ready equipment assets.}
     Given egocentric videos of human laboratory operations or multi-view videos of robot demonstrations, \method reconstructs laboratory equipment as simulation-ready assets with realistic geometry, materials, and articulated physical properties. The resulting assets can be directly integrated into SAPIEN to build interactive simulation environments for robot training, supporting diverse laboratory manipulation tasks such as object handling, liquid pouring, and equipment operation.}
\end{center}
}]
\begin{abstract}
Generating laboratory assets that faithfully reproduce the appearance, structure, and joint motions of real equipment remains challenging. Existing asset generation methods struggle to directly generate simulation assets for all equipment observed in operation videos and to faithfully reproduce the appearance, joint motions, and internal cavities of the real equipment. To address the challenge, we present \method, an agentic framework that generates editable, simulation-ready laboratory equipment assets from egocentric or multi-view operation videos. \method comprises four specialized agents. The Evidence Agent combines video observations with laboratory knowledge to specify equipment parts, materials, and functional requirements, which guide the Asset Agent in constructing editable Blender assets. The independent Verifier checks the generated geometry and appearance against the evidence and requirements, providing targeted feedback for revision. The Simulation Integration Agent converts verified Blender assets into simulation instances with physical properties and articulated mechanisms, using execution feedback to refine their behavior. Experiments on human and robot laboratory scenes show that \method achieves higher visual fidelity and stronger interaction capability than coding-agent and real-to-sim baselines, including a Codex baseline using the same underlying model. We further demonstrate valve-controlled drainage and use the generated asset library to reproduce laboratory scenes and operations in RoboTwin, supporting reusable simulation environments for robotic laboratory tasks.
\end{abstract}

\section{Introduction}
\label{sec:intro}

Robotic automation is increasingly enabling scientific experimentation~\cite{burger2020mobile, szymanski2023alab, darvish2024organa}. Simulation provides an important environment for developing and evaluating these robotic systems, supporting policy training, demonstration generation, and task development before deployment in real laboratories~\cite{mu2025robotwin,chen2025robotwin2}. These applications rely on simulation assets that faithfully represent laboratory equipment in appearance, structure, physical properties, and mechanical interactions. The need for such assets is particularly acute in scientific laboratories~\cite{burger2020mobile,darvish2024organa,li2025labutopia}, where real-world experiments consume reagents, may expose operators to hazardous substances, and require repeated setup and resetting. Simulation can reduce these costs and risks, but its effectiveness depends on interactive assets that faithfully represent laboratory equipment.

Existing approaches to generating such assets can be broadly grouped into three categories. Laboratory simulators~\cite{li2025labutopia,li2024chemistry3d,lan2025autobio} typically rely on manually modeled or curated equipment libraries, with assets prepared for specific environments and benchmarks. Articulated reconstruction recovers object geometry and joints from observed states or part motions~\cite{jiang2022ditto,liu2023paris,liu2025videoartgs, kerr2024rsrd, peng2026itaco}. Real-to-sim pipelines convert images, videos, or demonstrations into interactive assets and environments~\cite{chen2024urdformer,mandi2024real2code,le2024articulateanything,torne2024rialto,dai2024acdc,ren2026real2gym}. However, these approaches face distinct limitations when applied to laboratory operation videos. Specifically, although existing methods have achieved promising results on everyday objects, we observe substantial shortcomings in the laboratory equipment assets they generate. These include inaccurate material appearance, incomplete internal structures, and inoperable articulated components such as stopcocks and screw caps.

The limitations arise from two aspects.  First, video-based methods often rely on object-centric videos that provide clear and sufficiently complete observations of the target object’s motion. This reliance limits their applicability to laboratory operation videos, which capture multi-object experimental workflows rather than systematically demonstrate each apparatus and its mechanisms. Second, modeling pipelines that prioritize external shape often lack targeted inspection of local material and structural details, as well as explicit construction of the cavities and connected internal passages required for laboratory operations. For example, reproducing a separatory funnel’s exterior and assigning a rotational joint to its stopcock does not establish the internal channel through which the valve controls drainage. Addressing these bottlenecks requires extracting equipment-specific evidence from multi-object operation videos and translating functional requirements into construction constraints on materials, internal geometry, and mechanical relationships.

To address the challenge, we propose \method, an agentic framework that generates editable laboratory equipment assets and their simulation instances from operation videos, as illustrated in Fig.~\ref{fig:teaser}. The framework comprises four specialized agents. The Evidence Agent combines video observations with laboratory domain knowledge to establish the structure, materials, and functional requirements of each apparatus, which guide the Asset Agent in constructing editable Blender assets and exporting their meshes. An independent Verifier checks the generated assets against the evidence and requirements and guides revisions, while Simulation Integration converts verified assets into interactive simulation instances with physical properties and articulated mechanisms. A Coordinator orchestrates the workflow and routes tasks and feedback among these modules.

We evaluate \method on a human-performed chemistry workflow and a multi-camera robot scene, testing required part motions and valve-controlled drainage in simulation. Using the same model and tools as the Codex baseline, \method improves the parts-and-articulation score from 66.4 to 86.9 on the human workflow and from 76.8 to 98.3 on the robot scene. All required actions are successfully executed in the robot scene, and assets from both settings can be loaded and actuated natively in SAPIEN.

Our contributions are as follows: 
\begin{itemize}
    \item We present \method, an agentic framework that addresses the automatic generation of editable, simulation-ready laboratory equipment assets from multi-object operation videos.
    \item  We design four specialized agents—the Evidence Agent, Asset Agent, Verifier, and Simulation Integration Agent—orchestrated by a Coordinator to extract equipment requirements, construct Blender assets, perform independent verification, and generate interactive simulation instances in SAPIEN. 
    \item  Experiments on a human-performed chemistry workflow and a multi-camera robot scene demonstrate that \method produces more complete and kinematically accurate equipment assets than a Codex baseline using the same model and tools.
\end{itemize}

\section{Related Work}
\label{sec:related}

\noindent\textbf{Embodied intelligence for science.}
Robotic platforms support automated chemical synthesis and materials experimentation~\cite{steiner2019chemputer,burger2020mobile,szymanski2023alab,abolhasani2023sdl,tom2024sdl}, while language models support experimental planning, tool use, and execution control~\cite{boiko2023coscientist,bran2023chemcrow,skreta2023clairify,darvish2024organa}. Complementing these physical systems, laboratory simulators provide environments for training and evaluating robotic manipulation. LabUtopia~\cite{li2025labutopia} integrates high-fidelity laboratory scenes and instrument assets with a hierarchical task benchmark; Chemistry3D~\cite{li2024chemistry3d} supports the simulation of liquids and transparent objects; and AutoBio~\cite{lan2025autobio} combines instrument digitization with physics plugins for laboratory mechanisms. Extending these simulation resources to equipment beyond their existing libraries requires assets that reproduce the apparatus’s appearance, structure, and operating mechanisms. \method addresses this asset-creation need by generating editable equipment models and interactive simulation instances from laboratory operation videos.

\noindent\textbf{Multi-agent systems.}
LLM-based multi-agent systems decompose complex tasks among specialized agents that collaborate through role-playing, structured workflows, or debate~\cite{li2023camel,wu2024autogen,hong2024metagpt,qian2024chatdev,du2024debate}. In 3D content creation, 3D-GPT~\cite{sun2023threedgpt} coordinates specialized agents for procedural modeling, SceneCraft~\cite{hu2024scenecraft} synthesizes scenes as Blender scripts, and BlenderAlchemy~\cite{huang2024blenderalchemy} uses vision-language models to guide scene editing. Closely related to our asset construction, LL3M~\cite{lu2025ll3m} generates editable Blender code through multi-agent collaboration and refines assets through visual critique and verification. BlenderGym~\cite{gu2025blendergym} benchmarks vision-language-model systems on code-based 3D editing tasks.
Building on multi-agent collaboration and programmatic 3D modeling, \method applies specialized agents to laboratory real-to-sim asset generation. The coordinated workflow transforms evidence from operation videos into editable equipment assets and interactive simulation instances.

\noindent\textbf{Asset generation.}
Reconstruction methods recover geometry from observed views~\cite{mildenhall2020nerf,kerbl2023gaussian} and articulation from object states or motion~\cite{jiang2022ditto,liu2023paris,liu2025videoartgs}. Learned generators synthesize 3D assets~\cite{xiang2024trellis}, while programmatic methods construct geometry and articulation through rules or code~\cite{raistrick2023infinigen,mandi2024real2code,le2024articulateanything}. At the scene level, SimFoundry~\cite{ranawaka2026simfoundry} combines object extraction, mesh generation, and physical scene compilation to create interactive environments from videos. Real2Gym~\cite{ren2026real2gym} converts demonstrations into editable simulation environments with physics-based action validation. Our work focuses on generating equipment assets from multi-object laboratory operation videos. By combining video evidence with laboratory domain knowledge, \method models equipment appearance, internal structures, and functional mechanisms to produce editable assets and interactive simulation instances.
\section{Method}
\label{sec:method}

\begin{figure*}[t]
  \centering
  \includegraphics[
    width=\textwidth,
    trim={0mm 0mm 0mm 0mm},
    clip
  ]{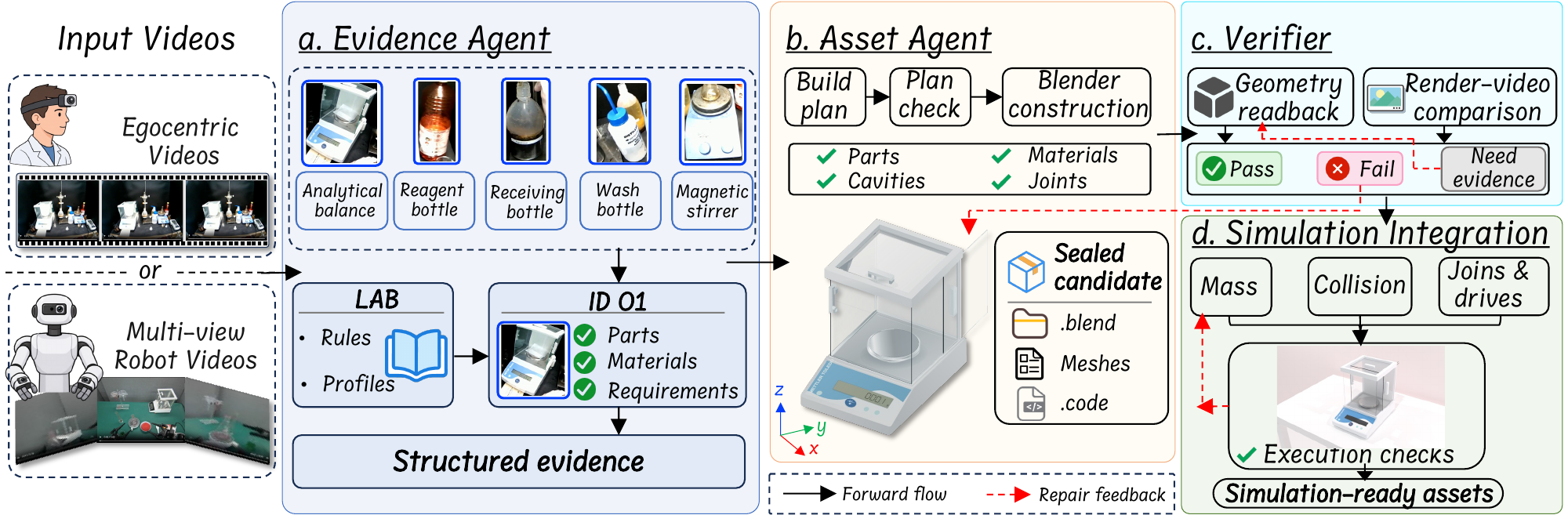}
  \caption{\textbf{Overview of \method.}
    Given egocentric or robot-recorded operation videos $\mathcal{V}$,
    \textbf{(a) the Evidence Agent} organizes equipment instances,
    their components, visual references, and physical information
    into structured evidence.
    \textbf{(b) The Asset Agent} translates this evidence into a
    build plan and constructs Blender source assets with exported
    meshes.
    \textbf{(c) An independent Verifier} inspects candidate geometry
    and materials and compares renders with video references,
    approving assets for simulation integration or specifying
    revisions for the Asset Agent.
    \textbf{(d) Simulation Integration} derives physical properties,
    collision shapes, and motion constraints from verified, sealed
    source assets to generate simulation instances.
    A Coordinator routes tasks and feedback; red dashed arrows
    indicate asset revision and simulation adaptation.
  }
  \label{fig:pipeline}
\end{figure*}

\method generates reusable laboratory equipment assets from operation
videos by combining video observations with equipment-specific
knowledge, as shown in Fig.~\ref{fig:pipeline}.
The resulting structural, material, and functional specifications
guide the construction of editable Blender source assets.
Independent verification relates the constructed assets to these
specifications and their supporting observations, while simulation
integration realizes their physical properties and articulated behavior.

\subsection{Problem Formulation}
\label{sec:overview}

Given operation videos $\mathcal{V}=\{v_1,\dots,v_N\}$ captured from
one or more viewpoints, our goal is to generate an asset for each
piece of laboratory equipment and each relevant consumable.
The components of an apparatus are represented within its asset.
For each target, \method produces an editable Blender source
package $S_i$ and a corresponding simulation asset
\begin{equation}
  A_i = (\mathcal{G}_i, \mathcal{M}_i, \mathcal{J}_i,
         \mathcal{C}_i, \mathcal{P}_i),
  \label{eq:asset}
\end{equation}
where $\mathcal{G}_i$ denotes part-level geometry,
$\mathcal{M}_i$ materials, $\mathcal{J}_i$ joints and motion
constraints, $\mathcal{C}_i$ collision shapes, and $\mathcal{P}_i$
mass, center of mass, and inertia.
The asset $A_i$ can be instantiated in
SAPIEN~\cite{xiang2020sapien}.

The generated representation should capture the appearance,
structure, and operation of the target apparatus.
Its rendered appearance should agree with the source frames,
its geometry should contain the apparatus's parts, openings, and
cavities, and its joints and constraints should reproduce the
required part motions, including mechanisms whose operation is
not demonstrated in the videos.
We also require an editable source representation in which
individual parts and connections can be modified through
localized changes to the construction program.

\subsection{Agentic Asset Generation}
\label{sec:agentic}

To generate high-fidelity laboratory assets that support physical interaction in simulation, we develop a multi-agent framework comprising four modules: the Evidence Agent, the Asset Agent, the Verifier, and Simulation Integration.
Each operates in a separate session of a multimodal language
model with role-specific inputs and tools.
A deterministic Coordinator schedules generation asset by asset
and routes tasks and feedback among the modules.
Structured evidence, build plans, source packages, and verification
results provide the interfaces between stages, as detailed in
Supp.~\ref{app:coord}.

\noindent\textbf{Evidence Agent.}\label{sec:evidence}
Generating an asset collection from a multi-object operation video requires identifying distinct equipment instances, establishing their part relationships, and observing how their components move during use. The Evidence Agent therefore builds an equipment registry that links localized observations to stable instances: apparatus, apparatus-owned components, and consumables. Frames sampled over the full duration of each video reveal parts visible at different times, while additional cameras contribute complementary views. Across temporally ordered frames, the agent observes component motion relative to the apparatus and the associated state changes, linking demonstrated operations to the parts involved. Each observation retains its camera, timestamp, and pixel region, grounding both structural and motion evidence in the source videos.

To consolidate this evidence, the agent associates observations across frames and cameras and requests additional frames or crops when their correspondence is ambiguous. Unresolved associations are retained as explicit hypotheses. Each registered instance is annotated with its category, owned parts, materials, demonstrated operations and associated part motions, and functional requirements, including openings, cavities, valves, and joints. New part-level findings can reopen earlier association hypotheses, allowing equipment interpretation and instance association to inform one another. Each annotation records its supporting observations and whether it is directly observed or inferred from domain knowledge.

Reusable assets require descriptions of equipment structure and
function that extend beyond the particular views and actions
captured in a recording.
We complement the observations with three forms of
laboratory-equipment knowledge.
\emph{Category rules} specify the morphology and required
components of each apparatus family; a separatory funnel, for
example, is characterized by its reservoir and lower stopcock.
\emph{Material profiles} represent each part's material as
$m=(p,\mathcal{A})$.
The profile $p$ selects a substrate from a catalog with density,
optical parameters, and contact parameters drawn from manufacturer
references, while $\mathcal{A}$ describes the tint and finish of
visible regions such as coatings, ink, and labels.
This representation carries physical material properties and
region-level appearance into the subsequent construction stages.
A \emph{transparent-glass policy} assigns eligible colorless glass
a neutral substrate, treating the background and contents visible
through it as observation context.
The registry, reference frames, functional requirements, and resolved
materials form the structured evidence bundle used for asset
construction; Supp.~\ref{app:evidence} details its organization
and knowledge rules.

\noindent\textbf{Asset Agent.}\label{sec:asset}
To translate the structural, material, and motion evidence into
an editable 3D asset, the Asset Agent develops a build plan
for each target:
\begin{equation}
  \mathcal{B}_i = (\mathcal{N}_i, \mathcal{E}_i),
  \label{eq:build_plan}
\end{equation}
where $\mathcal{N}_i$ contains named parts with geometric and
material specifications, and $\mathcal{E}_i$ describes their
connections through joint types, axes, and limits.
Observed part motions and equipment-specific functional
requirements guide how these parts are assembled and move.

Guided by $\mathcal{B}_i$, the LLM writes modeling code and invokes
Blender tools through the Model Context Protocol (MCP) to construct
the asset.
Rendered views guide refinements of geometry and appearance,
while named parts and connections in the construction program
support localized edits.
The resulting Blender file, exported meshes, and construction
code form the source package $S_i$, which is sealed and passed
to the independent Verifier.
Construction details are provided in Supp.~\ref{app:asset}.

\noindent\textbf{Verifier.}\label{sec:verify}
Fidelity to the target equipment concerns both its rendered
appearance and the geometry and material assignments stored
in the generated files.
The Verifier therefore evaluates $S_i$ against the source video
evidence and construction requirements fixed before modeling.
These requirements and the build plan $\mathcal{B}_i$ define a
set of check targets $\mathcal{K}_i$, each associated with a
specific source item and, where applicable, an appearance region.
Geometric readback inspects part geometry, dimensions, mesh
validity, opening and cavity connectivity, and material bindings
in the sealed files.
Render-vs-video review compares requirement-derived renders
with the original frames, using close views along declared
openings and isolated views of articulated parts at joint limits.

These inspections assign each target $k\in\mathcal{K}_i$ an outcome
$y_i(k)\in\{\mathrm{Pass}, \mathrm{Fail}, \mathrm{Unresolved}\}$.
A candidate is accepted when all targets pass:
\begin{equation}
  \operatorname{accept}(S_i)
  =
  \bigwedge_{k\in\mathcal{K}_i}
  \big[y_i(k)=\mathrm{Pass}\big].
  \label{eq:gate}
\end{equation}
Accepted candidates become verified source assets and proceed
to Simulation Integration.
When the available views are insufficient to assess a target,
the Verifier generates additional renders or cross-sectional
views of the sealed candidate and reassesses the unresolved
checks using the resulting evidence.

For confirmed defects, the Verifier issues a revision request
identifying the failed targets, supporting images, and affected
parts or regions.
The Asset Agent revises $\mathcal{B}_i$ and updates the
construction program accordingly.
The resulting source package is sealed and submitted for
verification against all targets.
The verification and revision procedures are detailed in
Supp.~\ref{app:verify}.

\noindent\textbf{Simulation Integration Agent.}\label{sec:sim}
To support physical interaction and equipment operation,
Simulation Integration Agent converts each verified, sealed source
package $S_i$ into a simulation asset $A_i$ with mass properties,
collision shapes, simulator materials, and articulated motion.
The source geometry is first partitioned by material.
For partition $j$ of asset $i$, let $\rho_{ij}$ denote its
catalog density and $\mathcal{T}_{ij}$ the consistently oriented
boundary triangles of its watertight mesh.
Its mass is computed as
\begin{equation}
  M_{ij}
  =
  \frac{\rho_{ij}}{6}
  \sum_{f\in\mathcal{T}_{ij}}
  \mathbf{a}_f \cdot
  \left(\mathbf{b}_f \times \mathbf{c}_f\right),
  \label{eq:mass}
\end{equation}
where $\mathbf{a}_f$, $\mathbf{b}_f$, and $\mathbf{c}_f$ are the
vertices of triangle $f$.
The signed volume accounts for cavity boundaries, assigning
hollow vessels the mass of their material walls.
The center of mass and inertia are obtained from the same
signed tetrahedral decomposition, and parts joined by fixed
connections are aggregated into rigid bodies.

To realize contact interactions, the integration model uses
local geometry tools to inspect the source meshes and generate
and evaluate candidate collision shapes.
Static concave apparatus use triangle-mesh collision to preserve
their openings and cavities, while movable parts receive convex
collision shapes.
Material profiles provide the optical and contact parameters
of the simulator materials, carrying the material assignments
established during evidence construction into the simulation
representation.

Equipment-specific operations are realized by compiling the
movable connections in $\mathcal{E}_i$ into simulator joints
and motion constraints $\mathcal{J}_i$.
Revolute and prismatic joints represent mechanisms such as
rotating stopcocks and sliding draft-shield doors, with drive
gains derived from the mass and inertia of the moving subtree.
Buttons receive a prismatic joint and a return spring.
For screw closures, a screw constraint couples rotation $\theta$
with axial travel $z$:
\begin{equation}
  z-z_0 = \frac{\ell}{2\pi}(\theta-\theta_0),
  \label{eq:screw_motion}
\end{equation}
where $\ell$ is the thread lead and $(\theta_0,z_0)$ is the
reference configuration.
The constraint advances the cap along the neck as it turns
while engaged.
At disengagement, the cap becomes a free body and can re-engage
when realigned with the neck.
These mechanisms translate the planned part relationships
into executable equipment behaviors.

To prepare the generated assets for robotic laboratory tasks,
we instantiate them and execute their prescribed operations
in SAPIEN~\cite{xiang2020sapien}.
An independent execution check reads back material and collision
properties and joint states, and compares simulator renders
with the original references.
The integration model uses this feedback to select geometry
inspections and diagnostic executions and to refine collision
shapes, joint drives, and control settings.
Each refinement is evaluated under the same execution protocol
before being incorporated into the simulation representation.
The resulting asset $A_i$ is packaged with its verified source
asset $S_i$ and added to the asset library for reuse in laboratory
task scenes.
Physical conversion and adaptation details are provided in
Supp.~\ref{app:sim}.
\section{Experiments}
\label{sec:exp}

We compare \method with coding agents and two real-to-sim frameworks using the same input videos, evaluating instance completeness, visual fidelity, and functionality in simulation. Ablation studies assess the contributions of independent verification and repair, simulation integration, and video input.
\subsection{Experimental Setup}
\label{sec:setup}


\begin{table*}[t]
  \centering
  \small
  \caption{\textbf{Quantitative comparison.} A: instance IoU; B: visual fidelity; $C_{\mathrm{p}}$ / $C_{\mathrm{m}}$: part presence / action completion, each out of 50; $C=C_{\mathrm{p}}+C_{\mathrm{m}}$, summed before rounding; D: valve-controlled drainage (N/A on Scene~2, which contains no separatory funnel). A, B, C, and D are in $[0,100]$; higher is better; best results in bold.}
  \label{tab:main}
  \setlength{\tabcolsep}{4.5pt}
  
  \begin{tabular}{l cccccc cccccc}
    \toprule
    & \multicolumn{6}{c}{Scene 1 (Ego)}
    & \multicolumn{6}{c}{Scene 2 (Robot)} \\
    \cmidrule(lr){2-7}\cmidrule(lr){8-13}
    Method
      & A & B & $C_{\mathrm{p}}$ & $C_{\mathrm{m}}$ & $C$ & D
      & A & B & $C_{\mathrm{p}}$ & $C_{\mathrm{m}}$ & $C$ & D \\
    \midrule
    Real2Gym~\cite{ren2026real2gym}
      & 76.2 & 57.8 & 38.0 & \phantom{0}2.4 & 40.4 & 0
      & 90.9 & 53.8 & 37.9 & 21.4 & 59.4 & N/A \\
    SimFoundry~\cite{ranawaka2026simfoundry}
      & 47.8 & 32.4 & 19.4 & 0 & 19.4 & 0
      & 69.2 & 40.3 & 31.0 & 0 & 31.0 & N/A \\
    Codex + GPT-6-Astra (xhigh)
      & 90.0 & 69.6 & 44.4 & 22.0 & 66.4 & \textbf{100}
      & \textbf{100} & 73.5 & \textbf{48.3} & 28.6 & 76.8 & N/A \\
    Codex + GPT-6-Sol (xhigh)
      & 88.9 & 55.7 & 40.7 & \phantom{0}7.9 & 48.7 & 0
      & \textbf{100} & 61.6 & 43.1 & 21.4 & 64.5 & N/A \\
    \midrule
    \method (Ours)
      & \textbf{94.7} & \textbf{78.1} & \textbf{49.1}
      & \textbf{37.8} & \textbf{86.9} & \textbf{100}
      & \textbf{100} & \textbf{78.9} & \textbf{48.3}
      & \textbf{50.0} & \textbf{98.3} & N/A \\
    \bottomrule
  \end{tabular}
\end{table*}

\begin{figure*}[t]
  \centering
  \colorlet{qband}{cvprblue}
  \definecolor{qours}{RGB}{214,96,0}
  \setlength{\tabcolsep}{0.6pt}
  \setlength{\fboxsep}{0pt}
  \renewcommand{\arraystretch}{0}
  \newcommand{\qw}{0.1\textwidth}
  \newcommand{\qh}{0.084\textwidth}
  \newcommand{\qcell}[1]{{\setlength{\fboxrule}{0.3pt}\fcolorbox{gray!45}{white}{#1}}}
  \newcommand{\qours}[1]{{\setlength{\fboxrule}{0.6pt}\fcolorbox{qours}{white}{#1}}}
  \newcommand{\qmiss}{{\setlength{\fboxrule}{0.3pt}\fcolorbox{gray!45}{gray!10}{\parbox[b][\dimexpr\qh-0.504pt\relax][c]{\dimexpr\qw-0.6pt\relax}{\centering\scriptsize\color{gray}not\\recovered}}}}
  \newcommand{\qrow}[2][\qh]{\rotatebox{90}{\parbox{#1}{\centering\scriptsize #2}}}
  \newcommand{\qhead}[1]{\parbox[b]{\qw}{\centering\scriptsize #1}}
  \newcommand{\qscene}[3]{\multicolumn{#1}{@{}c@{}}{\colorbox{qband}{\makebox[#2]{\rule[-2.3pt]{0pt}{9.6pt}\small\color{white}\textbf{#3}}}}}
  \begin{tabular}{@{}c@{\hspace{2pt}}ccccc@{\hspace{3.5pt}}ccc@{}}
    & \qscene{5}{\dimexpr0.5\textwidth+4.8pt\relax}{Scene 1 (Ego)} & \qscene{3}{\dimexpr0.3\textwidth+2.4pt\relax}{Scene 2 (Robot)} \\[2.5pt]
    & \qhead{Balance} & \qhead{Separatory\\funnel} & \qhead{Wash\\bottle} & \qhead{Reagent\\bottle} & \qhead{Hotplate\\stirrer} & \qhead{Balance} & \qhead{Transfer\\funnel} & \qhead{Funnel\\stand} \\[2pt]
    \qrow[\qw]{Real world} & \includegraphics[width=\qw]{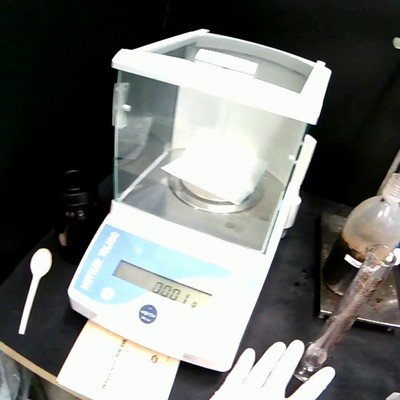} & \includegraphics[width=\qw]{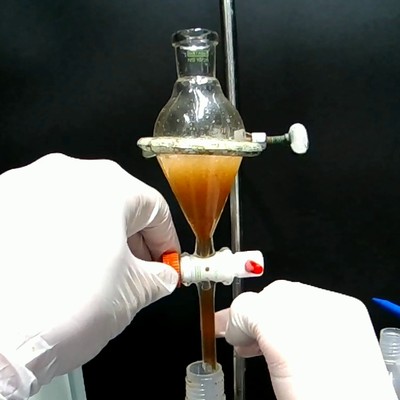} & \includegraphics[width=\qw]{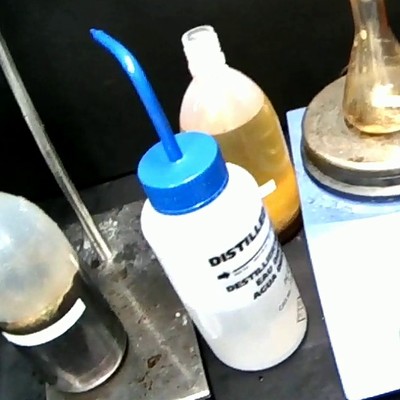} & \includegraphics[width=\qw]{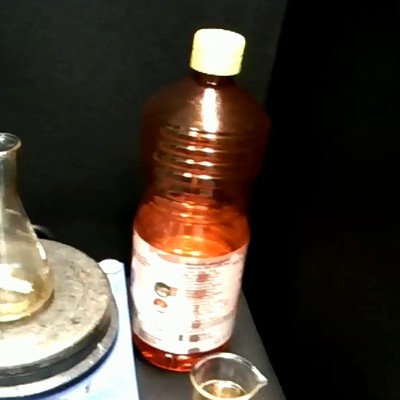} & \includegraphics[width=\qw]{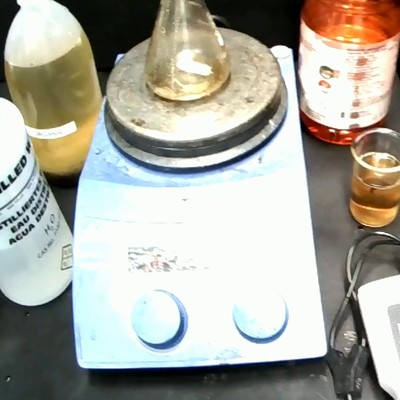} & \includegraphics[width=\qw]{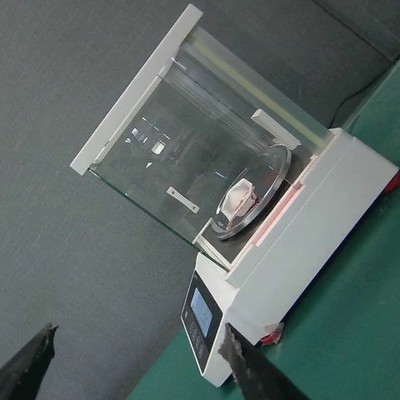} & \includegraphics[width=\qw]{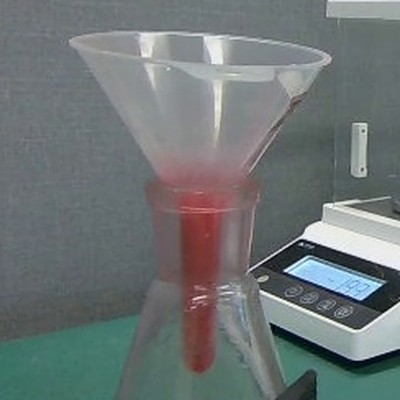} & \includegraphics[width=\qw]{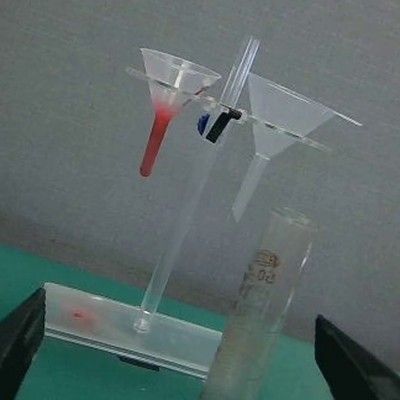} \\[1.2pt]
    \qrow{Real2Gym} & \qcell{\includegraphics[width=\dimexpr\qw-0.6pt\relax,trim=0 32 0 32,clip]{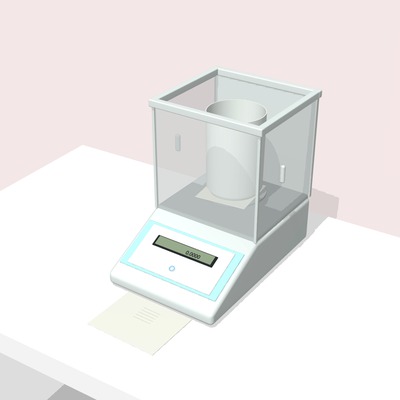}} & \qcell{\includegraphics[width=\dimexpr\qw-0.6pt\relax,trim=0 32 0 32,clip]{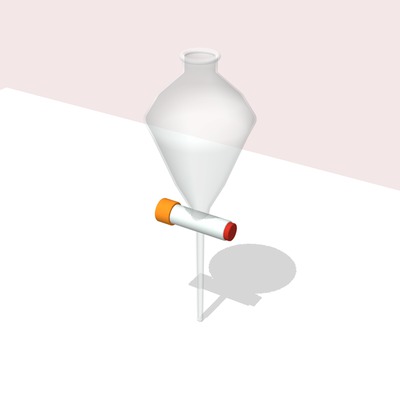}} & \qcell{\includegraphics[width=\dimexpr\qw-0.6pt\relax,trim=0 32 0 32,clip]{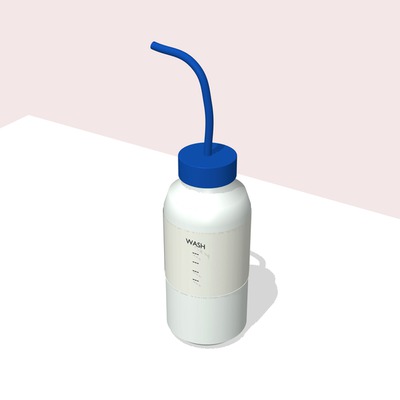}} & \qcell{\includegraphics[width=\dimexpr\qw-0.6pt\relax,trim=0 32 0 32,clip]{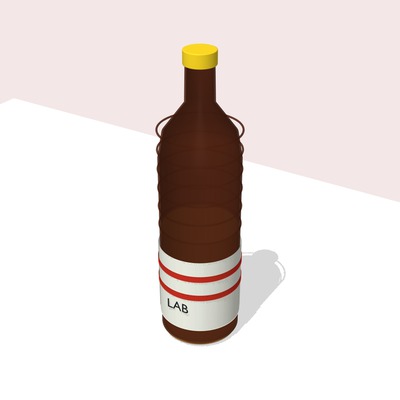}} & \qcell{\includegraphics[width=\dimexpr\qw-0.6pt\relax,trim=0 32 0 32,clip]{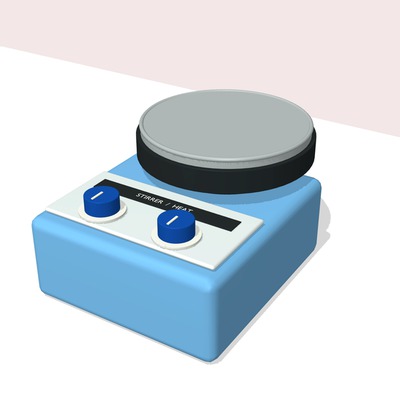}} & \qcell{\includegraphics[width=\dimexpr\qw-0.6pt\relax,trim=0 32 0 32,clip]{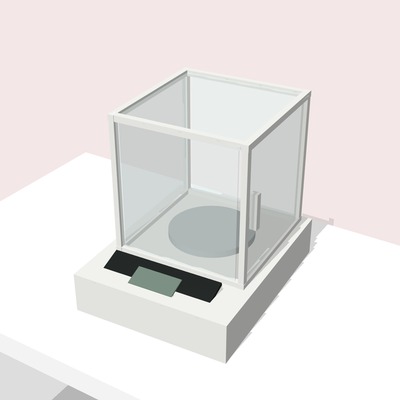}} & \qcell{\includegraphics[width=\dimexpr\qw-0.6pt\relax,trim=0 32 0 32,clip]{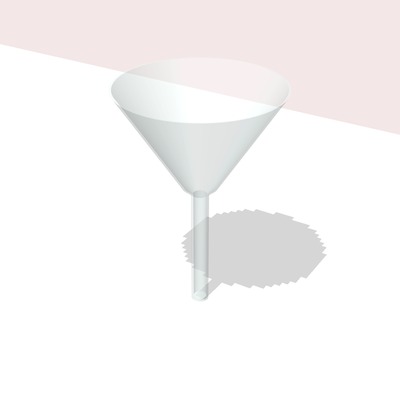}} & \qcell{\includegraphics[width=\dimexpr\qw-0.6pt\relax,trim=0 32 0 32,clip]{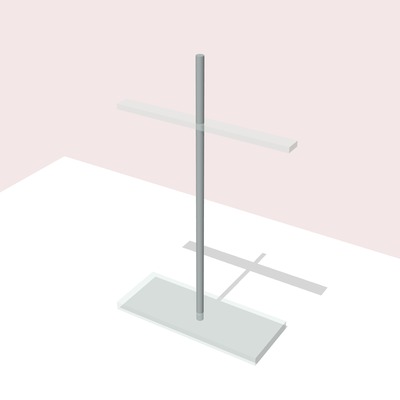}} \\[1.2pt]
    \qrow{\makebox[\linewidth][c]{SimFoundry}} & \qmiss & \qcell{\includegraphics[width=\dimexpr\qw-0.6pt\relax,trim=0 32 0 32,clip]{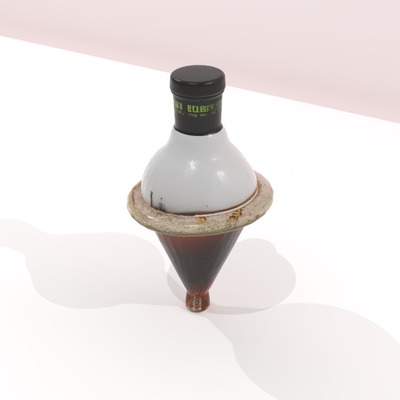}} & \qcell{\includegraphics[width=\dimexpr\qw-0.6pt\relax,trim=0 32 0 32,clip]{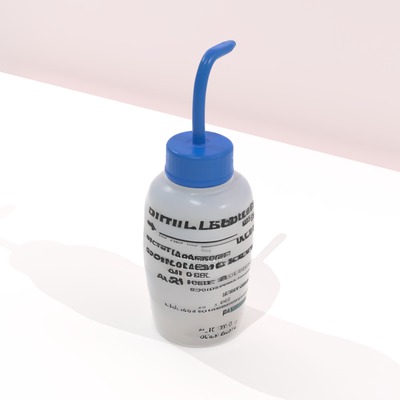}} & \qcell{\includegraphics[width=\dimexpr\qw-0.6pt\relax,trim=0 32 0 32,clip]{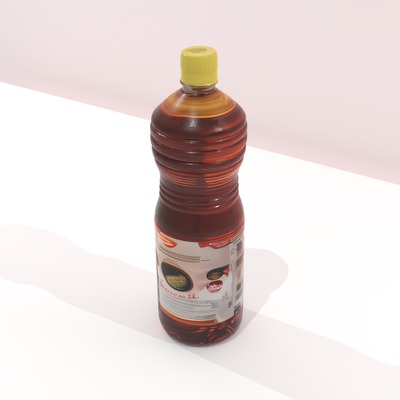}} & \qcell{\includegraphics[width=\dimexpr\qw-0.6pt\relax,trim=0 32 0 32,clip]{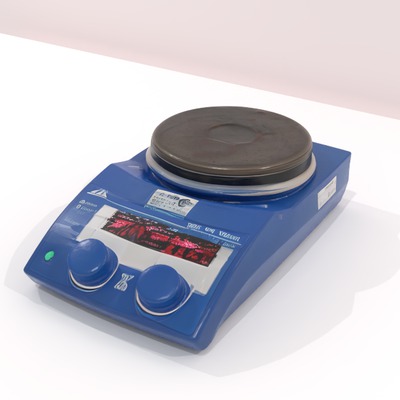}} & \qcell{\includegraphics[width=\dimexpr\qw-0.6pt\relax,trim=0 32 0 32,clip]{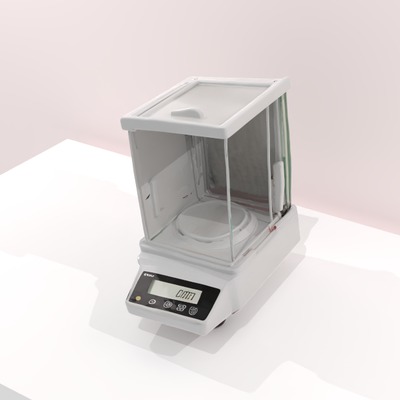}} & \qcell{\includegraphics[width=\dimexpr\qw-0.6pt\relax,trim=0 32 0 32,clip]{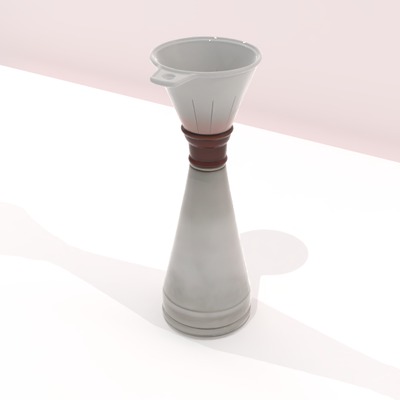}} & \qmiss \\[1.2pt]
    \qrow{Codex +\\\makebox[\linewidth][c]{GPT-6-Astra}} & \qcell{\includegraphics[width=\dimexpr\qw-0.6pt\relax,trim=0 32 0 32,clip]{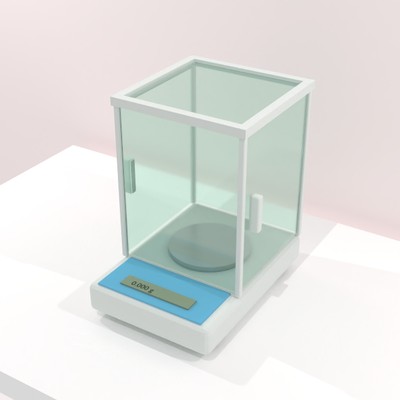}} & \qcell{\includegraphics[width=\dimexpr\qw-0.6pt\relax,trim=0 32 0 32,clip]{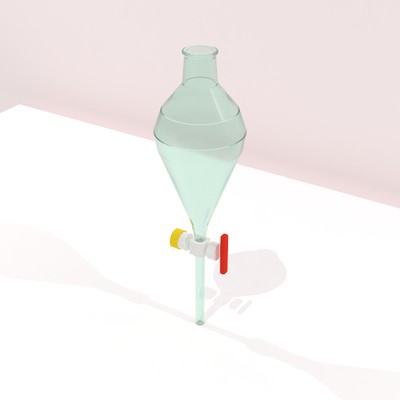}} & \qcell{\includegraphics[width=\dimexpr\qw-0.6pt\relax,trim=0 32 0 32,clip]{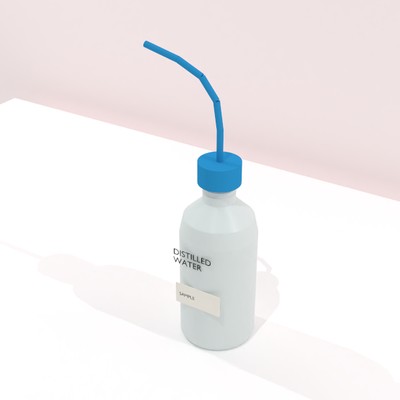}} & \qcell{\includegraphics[width=\dimexpr\qw-0.6pt\relax,trim=0 32 0 32,clip]{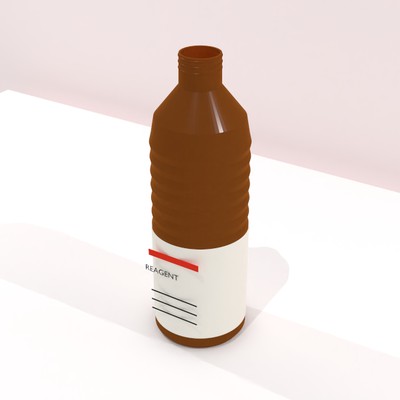}} & \qcell{\includegraphics[width=\dimexpr\qw-0.6pt\relax,trim=0 32 0 32,clip]{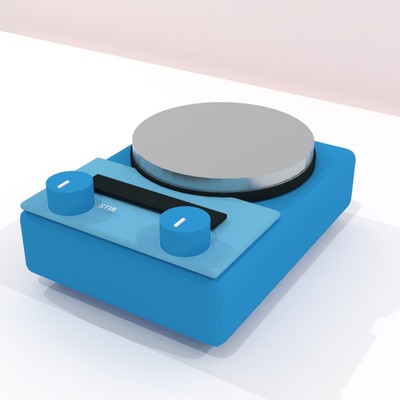}} & \qcell{\includegraphics[width=\dimexpr\qw-0.6pt\relax,trim=0 32 0 32,clip]{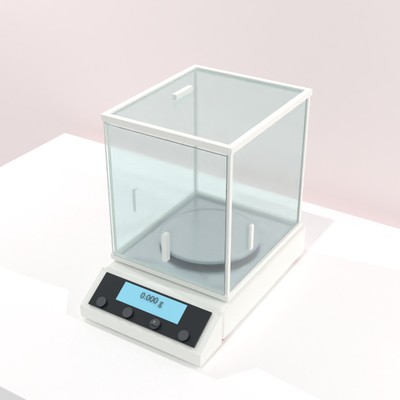}} & \qcell{\includegraphics[width=\dimexpr\qw-0.6pt\relax,trim=0 32 0 32,clip]{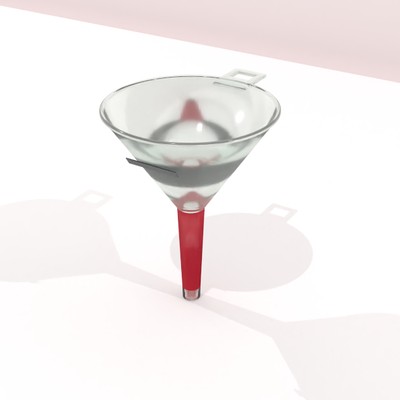}} & \qcell{\includegraphics[width=\dimexpr\qw-0.6pt\relax,trim=0 32 0 32,clip]{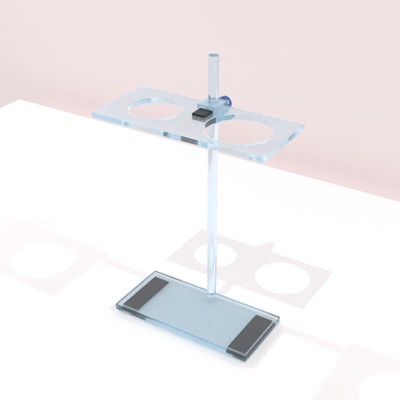}} \\[1.2pt]
    \qrow{Codex +\\GPT-6-Sol} & \qcell{\includegraphics[width=\dimexpr\qw-0.6pt\relax,trim=0 32 0 32,clip]{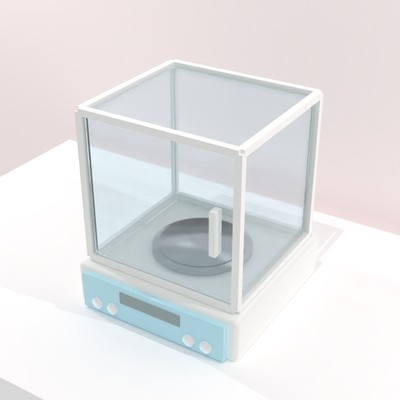}} & \qcell{\includegraphics[width=\dimexpr\qw-0.6pt\relax,trim=0 32 0 32,clip]{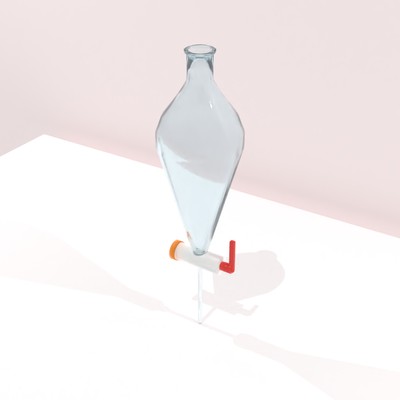}} & \qcell{\includegraphics[width=\dimexpr\qw-0.6pt\relax,trim=0 32 0 32,clip]{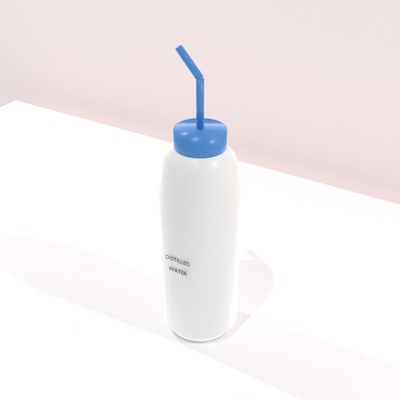}} & \qcell{\includegraphics[width=\dimexpr\qw-0.6pt\relax,trim=0 32 0 32,clip]{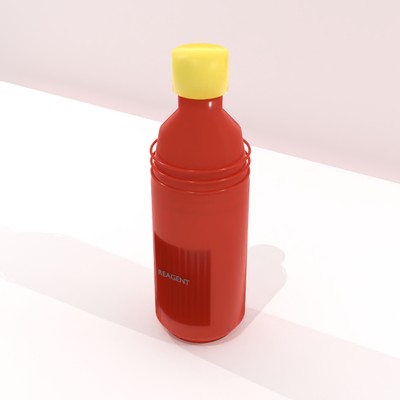}} & \qcell{\includegraphics[width=\dimexpr\qw-0.6pt\relax,trim=0 32 0 32,clip]{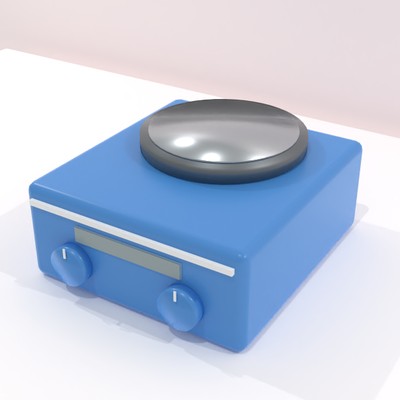}} & \qcell{\includegraphics[width=\dimexpr\qw-0.6pt\relax,trim=0 32 0 32,clip]{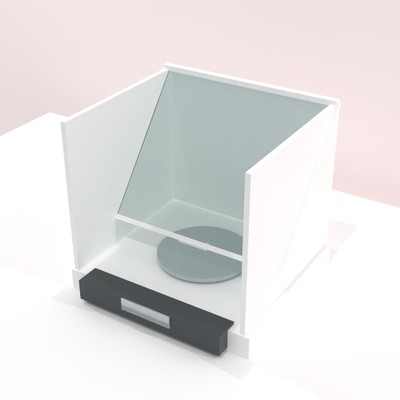}} & \qcell{\includegraphics[width=\dimexpr\qw-0.6pt\relax,trim=0 32 0 32,clip]{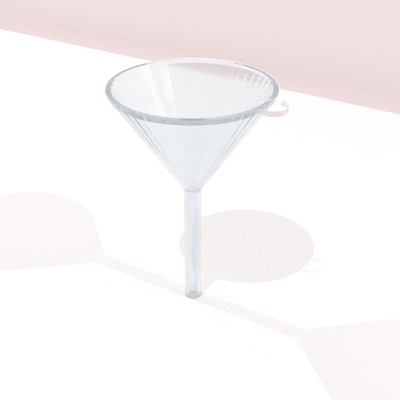}} & \qcell{\includegraphics[width=\dimexpr\qw-0.6pt\relax,trim=0 32 0 32,clip]{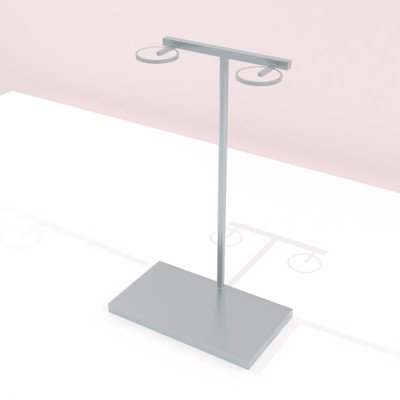}} \\[2.5pt]
    \qrow{\color{qours}\textbf{\method}\\\color{qours}(Ours)} & \qours{\includegraphics[width=\dimexpr\qw-1.2pt\relax,trim=0 32 0 32,clip]{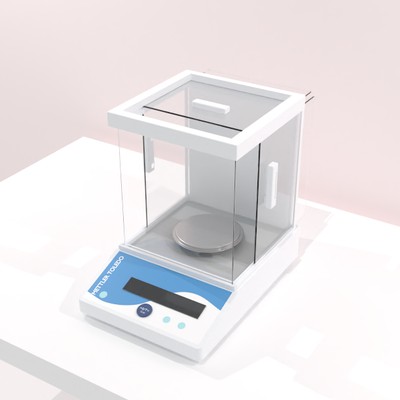}} & \qours{\includegraphics[width=\dimexpr\qw-1.2pt\relax,trim=0 32 0 32,clip]{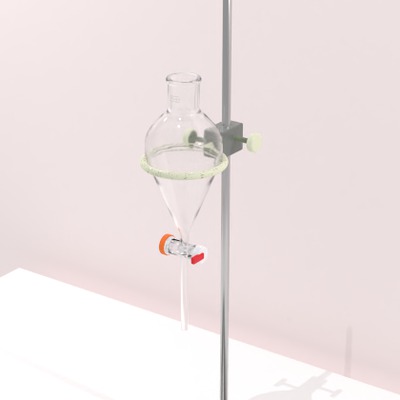}} & \qours{\includegraphics[width=\dimexpr\qw-1.2pt\relax,trim=0 32 0 32,clip]{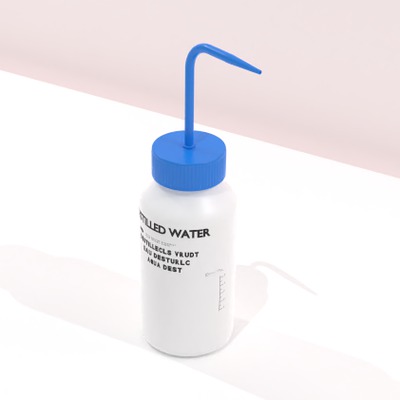}} & \qours{\includegraphics[width=\dimexpr\qw-1.2pt\relax,trim=0 32 0 32,clip]{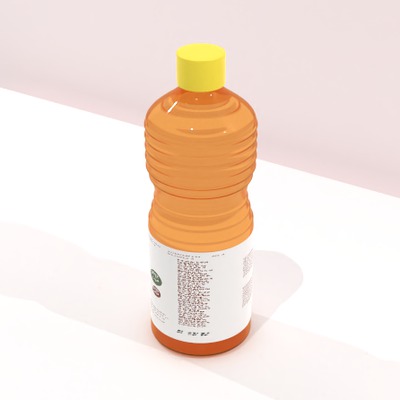}} & \qours{\includegraphics[width=\dimexpr\qw-1.2pt\relax,trim=0 32 0 32,clip]{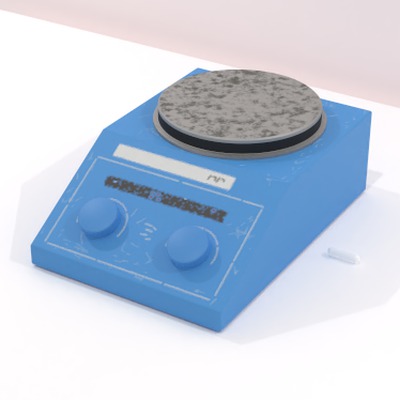}} & \qours{\includegraphics[width=\dimexpr\qw-1.2pt\relax,trim=0 32 0 32,clip]{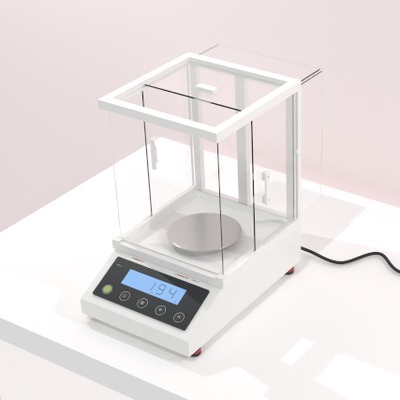}} & \qours{\includegraphics[width=\dimexpr\qw-1.2pt\relax,trim=0 32 0 32,clip]{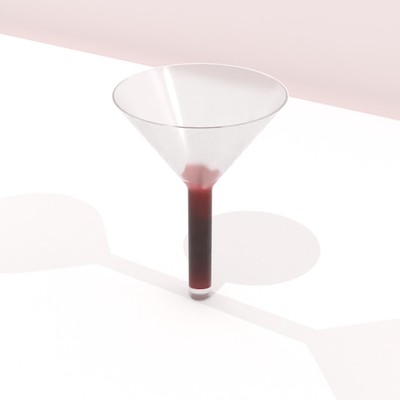}} & \qours{\includegraphics[width=\dimexpr\qw-1.2pt\relax,trim=0 32 0 32,clip]{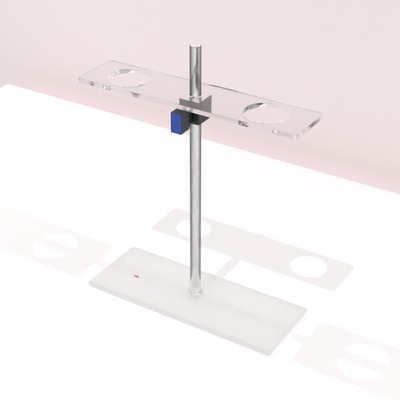}} \\
  \end{tabular}
  \caption{\textbf{Qualitative comparison.} The first row shows each instrument cropped from a source video frame; the remaining rows show the assets delivered by each method in the same tabletop scene with identical lighting and camera framing, rendered with SAPIEN ray tracing (Real2Gym's MuJoCo models are rendered in MuJoCo). Gray cells mark instruments that a method did not recover. More qualitative results are provided in Supp.~Fig.~\ref{fig:qual_full}.}
  \label{fig:qual}
\end{figure*}

\noindent\textbf{Scenes.}
We evaluate \method on two laboratory scenes covering human and robot workflows with different observation configurations.
Scene~1 comprises 57 egocentric clips totaling 228\,s from a public collection of laboratory-action videos~\cite{ibm2024labactions}, showing a human performing weighing, reagent addition, stirring, phase separation, layer collection, and pH measurement.
Scene~2 captures a robotic weighing-and-transfer task over 130\,s using three synchronized head and wrist cameras.
The scenes include transparent glassware, instruments with movable components such as draft-shield doors and screw caps, and assemblies such as a separatory funnel mounted on a ring stand.
Their reference annotations cover 29 equipment instances, 83 required parts, and 48 part--action pairs, spanning the appearance, structure, and operation of laboratory equipment.
Complete asset inventories are provided in Supp.~\ref{app:data}.

\noindent\textbf{Baselines.}
We compare with Codex driven by GPT-6-Astra or GPT-6-Sol at \emph{xhigh} reasoning effort, together with two real-to-sim frameworks, Real2Gym~\cite{ren2026real2gym} and SimFoundry~\cite{ranawaka2026simfoundry}.
Each Codex baseline receives the same source videos and Blender MCP server as \method and is tasked with producing editable Blender assets with visual meshes, collision geometry, mass properties, and joints for SAPIEN/RoboTwin.
The agents choose their own procedures for inspecting the videos, constructing the assets, and checking their outputs.
Real2Gym reconstructs editable simulation scenes from human and robot demonstrations.
SimFoundry combines object extraction, 3D mesh generation, and physical scene compilation in a modular video-to-simulation pipeline. We provide SimFoundry with the complete workflow video from the primary camera view.

\noindent\textbf{Evaluation metrics.}
To assess asset quality, we evaluate all methods along four complementary dimensions using shared reference annotations, with each dimension reported separately on a 0–100 scale. At the scene level, instance completeness (A) measures inventory recovery through IoU after one-to-one matching of reference and generated instances, penalizing missing, extra, and duplicated instances. For individual assets, visual fidelity (B) compares standardized renders with source frames to assess appearance, materials, and visible structure; missing instances receive zero.

We then assess interaction capability through parts and articulation (C), which sums required-part presence ($C_{\mathrm{p}}$) and action completion ($C_{\mathrm{m}}$), each scored out of 50. Actions are tested in simulation and include both demonstrated operations and those implied by visible structure. GPT-6-Astra with \emph{max} reasoning effort assesses visual fidelity and action completion under common scoring criteria.

To assess whether these interactions produce the intended physical outcome, valve-controlled drainage (D) reports the percentage of separatory funnels that retain liquid with the valve closed and drain through the lower outlet after opening without wall leakage. This test applies only to Scene~1. Detailed scoring and simulation protocols are provided in Supp.~\ref{app:protocol}.

\noindent\textbf{Implementation details.}
All agents of \method use GPT-6-Astra, with the Asset Agent operating at \emph{xhigh} reasoning effort, matching the corresponding Codex baseline.
We construct assets in Blender~5.2 and integrate them into RoboTwin on SAPIEN, allowing up to $T=5$ source-asset repairs per asset and bounded simulation adaptation.
Real2Gym's MuJoCo~\cite{todorov2012mujoco} outputs are evaluated natively in MuJoCo under the same action definitions.
Further implementation and engine-specific evaluation details are provided in Supp.~\ref{app:data} and Supp.~\ref{app:protocol}.

\subsection{Results}
\label{sec:results}

\begin{table}[t]
  \centering
  \small
  \caption{\textbf{Video input ablation.} \emph{images-input} replaces videos with still frames. Metrics follow Tab.~\ref{tab:main}.}
  \label{tab:ablation_image}
  \setlength{\tabcolsep}{3.5pt}
  \begin{tabular}{ll cccccc}
    \toprule
    Scene & Variant & A & B & $C_{\mathrm{p}}$ & $C_{\mathrm{m}}$ & $C$ & D \\
    \midrule
    \multirow{2}{*}{Scene 1}
      & \method (full) & 94.7 & 78.1 & 49.1 & 37.8 & 86.9 & 100 \\
      & images-input & 89.5 & 73.9 & 44.4 & 25.0 & 69.4 & 100 \\
    \midrule
    \multirow{2}{*}{Scene 2}
      & \method (full) & 100  & 78.9 & 48.3 & 50.0 & 98.3 & N/A \\
      & images-input & 90.9 & 75.9 & 48.3 & 50.0 & 98.3 & N/A \\
    \bottomrule
  \end{tabular}
\end{table}

\begin{table}[t]
  \centering
  \small
  \caption{\textbf{Feedback ablation.} \emph{w/o feedback} removes explicit verification feedback. Metrics follow Tab.~\ref{tab:main}.}
  \label{tab:ablation}
  \setlength{\tabcolsep}{3.5pt}
  \begin{tabular}{ll cccccc}
    \toprule
    Scene & Variant & A & B & $C_{\mathrm{p}}$ & $C_{\mathrm{m}}$ & $C$ & D \\
    \midrule
    \multirow{2}{*}{Scene 1}
      & \method (full) & 94.7 & 78.1 & 49.1 & 37.8 & 86.9 & 100 \\
      & w/o feedback & 94.7 & 73.4 & 49.1 & 31.1 & 80.2 & 0 \\
    \midrule
    \multirow{2}{*}{Scene 2}
      & \method (full) & 100  & 78.9 & 48.3 & 50.0 & 98.3 & N/A \\
      & w/o feedback & 100  & 74.5 & 46.6 & 42.9 & 89.4 & N/A \\
    \bottomrule
  \end{tabular}
\end{table}

\begin{table}[t]
  \centering
  \small
  \caption{\textbf{Simulation integration ablation.} \emph{w/o adaptation} removes model-guided adjustments and refinement; \emph{w/o joint presets} uses agent-authored joint configurations. Metrics follow Tab.~\ref{tab:main}.}
  \label{tab:ablation_sim}
  \setlength{\tabcolsep}{3.2pt}
  \begin{tabular}{ll cccccc}
    \toprule
    Scene & Variant & A & B & $C_{\mathrm{p}}$ & $C_{\mathrm{m}}$ & $C$ & D \\
    \midrule
    \multirow{3}{*}{Scene 1}
      & \method (full) & 94.7 & 78.1 & 49.1 & 37.8 & 86.9 & 100 \\
      & w/o adaptation & 94.7 & 78.1 & 49.1 & 32.9 & 82.0 & 0 \\
      & w/o joint presets & 94.7 & 78.1 & 49.1 & 32.9 & 82.0 & 0 \\
    \midrule
    \multirow{3}{*}{Scene 2}
      & \method (full) & 100  & 78.9 & 48.3 & 50.0 & 98.3 & N/A \\
      & w/o adaptation & 100  & 78.9 & 48.3 & 50.0 & 98.3 & N/A \\
      & w/o joint presets & 100  & 78.9 & 48.3 & 39.3 & 87.6 & N/A \\
    \bottomrule
  \end{tabular}
\end{table}
Tab.~\ref{tab:main} presents the experimental results, showing that \method recovers the complete reference equipment inventory in both scenes and achieves instance completeness (A) scores of 94.7 and 100 in Scene~1 and Scene~2, respectively. All 18 reference instances in Scene~1 and all 11 in Scene~2 are recovered, with one additional instance on Scene~1. This coverage spans the balances, glassware, and support structures used throughout the observed laboratory procedures, demonstrating consistent equipment recovery from both human and robot workflows.

\method also achieves the highest visual fidelity (B) in both scenes, scoring 78.1 and 78.9 compared with 69.6 and 73.5 for Codex + GPT-6-Astra. Fig.~\ref{fig:qual} illustrates its preservation of characteristic equipment geometry and layout, including the framed draft shields and weighing platforms of the balances, the printed wash bottle markings, and the separatory funnel with its support stand. Compared with the viewpoint coverage recommended by SimFoundry, our workflow videos cover a narrower range of viewing angles, which may partly explain the reconstruction differences. \method{}’s results show that faithful equipment structures can be recovered from these videos of laboratory operations.

The strongest improvements are observed in parts and articulation (C), where \method reaches 86.9 on Scene~1 and 98.3 on Scene~2, compared with 66.4 and 76.8 for Codex + GPT-6-Astra. The gains are primarily driven by action completion: \(C_{\mathrm{m}}\) increases from 22.0 to 37.8 on Scene~1, while all seven required actions are completed on Scene~2. These results demonstrate that \method translates the equipment’s required mechanisms into executable interactions more effectively, connecting the recovered component structure with the motions needed to operate the equipment.

\method further achieves 100 in the Scene~1 valve-controlled drainage test (D), matching Codex + GPT-6-Astra. Its separatory funnel retains liquid with the valve closed and drains through the lower outlet after opening, showing that the vessel, valve, and internal flow path jointly support the intended function. The usable internal cavity recovered by \method is essential to this behavior; its absence in the SimFoundry funnel prevents liquid initialization. The combination of complete equipment coverage, high visual fidelity, improved action completion, and successful drainage demonstrates the effectiveness of our collaborative multi-agent framework in generating interactive laboratory assets.

\noindent\textbf{Ablation studies.}
We examine how video evidence, explicit verification feedback, and simulation integration contribute to asset fidelity and operability. The \emph{images-input} variant in Tab.~\ref{tab:ablation_image} replaces videos with 25 selected frames from Scene~1 and three from each of Scene~2's three cameras, keeping downstream components unchanged. Instance completeness and visual fidelity decline in both scenes, while reduced part presence and action completion lower Scene~1's interaction score from 86.9 to 69.4. Scene~2 retains its interaction performance but misses the folded weighing paper. These results indicate that video provides more complete observations and temporal context for recovering equipment structure and motion, supporting the generation of more faithful and operable assets.

Beyond the input evidence, Tab.~\ref{tab:ablation} examines the contribution of explicit verification feedback. The \emph{w/o feedback} variant retains the Asset Agent's initial submissions without Verifier-guided correction or subsequent repair. Visual fidelity falls to 73.4 and 74.5 in Scenes~1 and~2, respectively, and part presence decreases in Scene~2. Interaction performance also declines, and the Scene~1 funnel retains liquid when closed but fails to drain after opening. Explicit feedback and subsequent repair therefore enable targeted corrections to the initial assets, improving their appearance, structural completeness, and operability.

To isolate simulation integration, the ablations in Tab.~\ref{tab:ablation_sim} reuse the full framework's Blender assets, leaving instance completeness, visual fidelity, and part presence unchanged. Both variants produce assets with misaligned joints, and their Scene~1 funnels fail the drainage test because of lateral leakage. The \emph{w/o adaptation} variant removes the integration model's initial configuration adjustments and subsequent feedback-driven refinement, reducing Scene~1's interaction score to 82.0 while leaving Scene~2 unchanged. These results indicate that model-guided adaptation supports the construction and refinement of collision geometry and joint configurations, helping the integrated assets realize their intended physical behavior.

The \emph{w/o joint presets} variant replaces prescribed motion rules with agent-authored native joint configurations. Action completion decreases from 37.8 to 32.9 in Scene~1 and from 50.0 to 39.3 in Scene~2. Joint-motion presets thus provide explicit constraints on how the generated mechanisms move, enabling interactions that more closely match the operation of real laboratory equipment.
\begin{figure}[t]
  \centering
  \setlength{\tabcolsep}{0.8pt}
  \renewcommand{\arraystretch}{0.8}
  \newcommand{\rw}{0.23\linewidth}
  \newcommand{\rrow}[1]{\rotatebox{90}{\parbox{0.1725\linewidth}{\centering\scriptsize #1}}}
  \newcommand{\rhead}[1]{\parbox[b]{\rw}{\centering\scriptsize #1}}
  \begin{tabular}{ccccc}
    & \rhead{Open shield} & \rhead{Press key} & \rhead{Pick up funnel} & \rhead{Pour liquid} \\
    \rrow{Real} & \includegraphics[width=\rw]{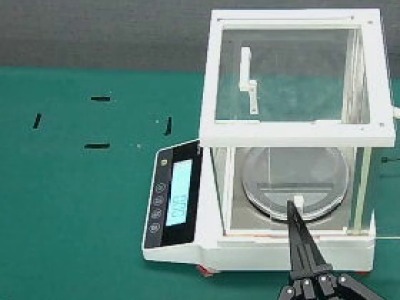} & \includegraphics[width=\rw]{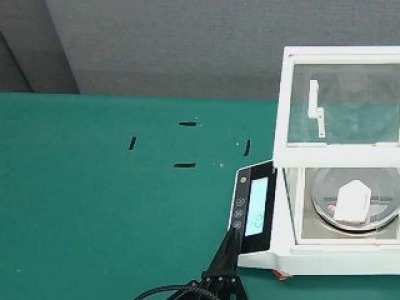} & \includegraphics[width=\rw]{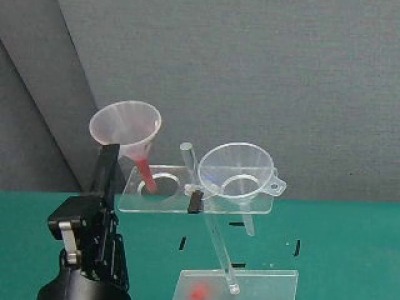} & \includegraphics[width=\rw]{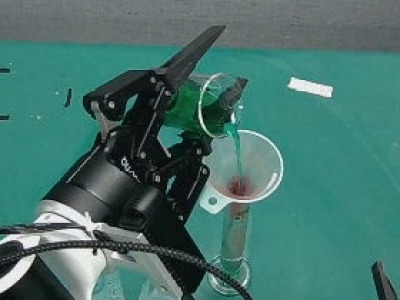} \\
    \rrow{Sim} & \includegraphics[width=\rw]{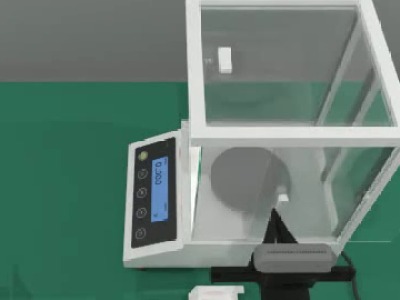} & \includegraphics[width=\rw]{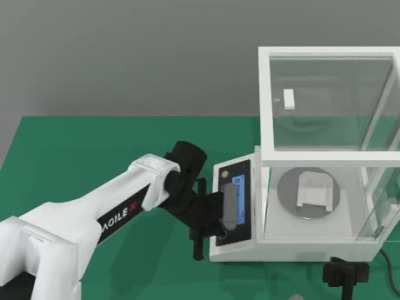} & \includegraphics[width=\rw]{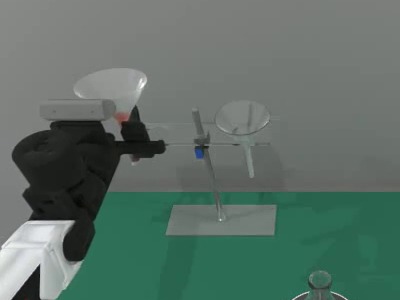} & \includegraphics[width=\rw]{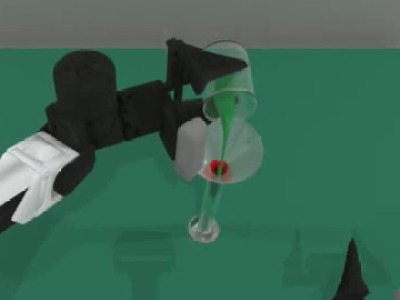} \\
  \end{tabular}
  \caption{\textbf{Laboratory manipulation in RoboTwin.} Given real operation videos and the asset library generated by \method, GPT-6-Astra reconstructs the observed scenes and reproduces the demonstrated operations. Top: real-world head-camera frames; bottom: simulated reproductions at corresponding stages. Four representative operations are shown, with examples from all five task episodes provided in Supp.~Fig.~\ref{fig:real2sim}.}
  \label{fig:tasks}
\end{figure}

\noindent\textbf{Laboratory manipulation in RoboTwin.}
To demonstrate the use of the generated assets in laboratory workflows, we provide GPT-6-Astra with videos of real laboratory operations and access to the asset library produced by \method. GPT-6-Astra uses these assets to reconstruct the observed equipment layouts in RoboTwin and reproduce the demonstrated operations. Fig.~\ref{fig:tasks} compares corresponding stages of real and simulated executions for four representative operations: opening a balance’s draft-shield door, pressing a button, picking up a funnel, and pouring liquid. These demonstrations bring the generated assets into laboratory task environments that support equipment operation, object manipulation, and liquid handling, illustrating how the asset library can be used to construct interactive simulation tasks from video demonstrations.

\section{Conclusion}
\label{sec:conclusion}

We presented \method, a closed-loop multi-agent framework that generates editable, simulation-ready equipment assets from videos of chemistry experiments.
By separating construction from independent verification and incorporating simulation feedback, the framework aligns asset appearance and structure with video evidence while validating the mechanisms required for equipment operation. Experiments on egocentric human demonstrations and multi-camera robot recordings show higher visual fidelity and stronger interaction capability than the compared baselines. Ablations further demonstrate the  contributions of closed-loop feedback, simulation integration, and video observations to asset quality and usability. Using the generated asset library, GPT-6-Astra reconstructs observed scenes and reproduces demonstrated operations in RoboTwin, illustrating the use of these assets for equipment operation, object manipulation, and liquid handling. These results connect video-based asset generation to the construction of reusable, interactive laboratory environments for robotic tasks.

\section{LLM Usage Statement}

During the preparation of this manuscript and the development of the proposed framework, we used large language models (LLMs) and generative AI tools for several purposes.

GPT-6-Astra served as the underlying model for the Evidence Agent, Asset Agent, Verifier, and Simulation Integration module in \method. It assisted with video evidence interpretation, asset construction planning, tool orchestration, verification feedback analysis, and simulation asset adaptation. GPT-6-Astra was also used to evaluate generated assets. Given the evaluation criteria and scoring rubrics defined by the authors, it acted as a tool-using evaluation agent, invoking tools to inspect rendered appearances, examine geometric properties, test articulated-part motions, and assess functional behaviors in simulation. It assigned scores based on the evidence collected through these tool-assisted evaluations and the predefined criteria.

Claude Opus 5.5 and GPT-6-Astra were used to assist with experimental code development, including code generation, implementation, debugging, and refinement. For baseline experiments, Codex driven by GPT-6-Astra and GPT-6-Sol was used as a coding-agent baseline, as described in the experimental setup. ChatGPT was also used solely for grammar correction and language polishing of the manuscript.

In addition, ChatGPT's image generation capability was used to assist in creating the \method icon in the teaser figure. The generated icon was subsequently edited and incorporated into the final figure by the authors.

All research ideas, methodologies, evaluation metrics, scoring criteria, and experimental protocols were developed and defined by the authors. The authors reviewed and verified the generated code, assets, tool-assisted evaluation results, and all technical content. The authors take full responsibility for the accuracy, originality, and completeness of the submitted work.
{
    \small
    \bibliographystyle{ieeenat_fullname_unsrt}
    \bibliography{main}
}

\clearpage
\maketitlesupplementary
\appendix

\section{Method Details}
\label{app:method}

This section complements Sec.~\ref{sec:agentic} with the data structures, rules, and parameters of each stage of \method.

\subsection{Coordination}
\label{app:coord}

The Coordinator follows the deterministic schedule in
Alg.~\ref{alg:coordinator}.
It compiles the evidence bundle $\mathcal{D}$ into construction
tasks $\tau_i$ through $\operatorname{CompileTasks}$ and processes
them asset by asset.
For each task, the Asset Agent produces an admitted build plan
$\mathcal{B}_i^{(0)}$ and a sealed source candidate $S_i^{(0)}$.

At round $t$, the Verifier derives the check targets $\mathcal{K}_i$
from $\tau_i$ and $\mathcal{B}_i^{(t)}$, acquires targeted evidence
for unresolved targets, and returns the outcomes
$y_i=\{y_i(k)\}_{k\in\mathcal{K}_i}$ together with the requested
edits $F_i$, as described in Supp.~\ref{app:verify}.
The Coordinator routes $F_i$ to the Asset Agent, which updates
the plan and construction program and submits
$(\mathcal{B}_i^{(t+1)},S_i^{(t+1)})$ for re-verification.
The loop ends when the source candidate is accepted
according to Eq.~\eqref{eq:gate}, when $F_i=\emptyset$, or after $T$ repairs.

Accepted sources proceed to Simulation Integration.
After the repair budget is exhausted, the Coordinator may also
select an available sealed candidate for diagnostic import.
The Simulation Integration Agent integrates the selected source
$S_i$ into $A_i$ with adaptation budget $T_{\mathrm{sim}}$ and
returns the import-check report $e_i$, as described in Supp.~\ref{app:sim}.
Pairs $(S_i,A_i)$ that pass the import check enter $\mathcal{L}$
with their source verification outcomes.

\newcounter{suppalg}
\providecommand{\theHsuppalg}{\arabic{suppalg}}

\begin{figure}[t]
  \small
  \newcommand{\coordkw}[1]{\textbf{#1}}
  \newcommand{\coordagent}[1]{\textnormal{#1}}
  \newcommand{\coordbar}{%
    \makebox[1.15em][l]{\hspace*{.27em}%
      \raisebox{0pt}[0pt][0pt]{\vrule width .35pt
        height \dimexpr\ht\strutbox+1.4pt\relax
        depth \dimexpr\dp\strutbox+1.4pt\relax}}}
  \refstepcounter{suppalg}\label{alg:coordinator}%
  \hrule height .8pt\relax
  \kern3pt
  \noindent\textbf{Algorithm~\thesuppalg}\enspace
  Coordinator schedule of \method\par
  \kern3pt\hrule height .4pt\relax\kern3pt
  \noindent\coordkw{Input:} Operation videos \(\mathcal V=\{v_1,\ldots,v_N\}\),\newline
  source repair budget \(T\), and simulation adaptation budget \(T_{\mathrm{sim}}\).\par
  \noindent\coordkw{Output:} Asset library \(\mathcal L\) of source--simulation pairs \((S_i,A_i)\).\par

  \begin{enumerate}[
    label={\footnotesize\arabic*:},leftmargin=1.95em,labelwidth=1.5em,
    labelsep=.45em,align=right,itemsep=0pt,parsep=0pt,topsep=3pt,partopsep=0pt]
    \item \(\mathcal D\gets\coordagent{Evidence Agent}(\mathcal V)\)
    \item \((\tau_1,\ldots,\tau_n)\gets\operatorname{CompileTasks}(\mathcal D)\)
    \item \(\mathcal L\gets\emptyset\)
    \item \coordkw{for} \(i=1,\ldots,n\) \coordkw{do}
    \item \coordbar\((\mathcal B_i^{(0)},S_i^{(0)})\gets
      \coordagent{Asset Agent}.\mathrm{Build}(\tau_i)\)
    \item \coordbar\coordkw{for} \(t=0,\ldots,T\) \coordkw{do}
    \item \coordbar\coordbar\((y_i,F_i)\gets
      \coordagent{Verifier}(\tau_i,\mathcal B_i^{(t)},S_i^{(t)})\)
    \item \coordbar\coordbar\coordkw{if} \(\operatorname{accept}(S_i^{(t)})\)
      \coordkw{or} \(F_i=\emptyset\) \coordkw{or} \(t=T\) \coordkw{then break}
    \item \coordbar\coordbar\((\mathcal B_i^{(t+1)},S_i^{(t+1)})\gets{}\)\newline
      \coordbar\coordbar\hspace*{1.15em}%
      \(\coordagent{Asset Agent}.\mathrm{Revise}(\mathcal B_i^{(t)},S_i^{(t)},F_i)\)
    \item \coordbar\coordkw{end for}
    \item \coordbar\coordkw{if} a sealed source \(S_i\) is selected for import \coordkw{then}
    \item \coordbar\coordbar\((A_i,e_i)\gets
      \coordagent{Simulation Integration Agent}(S_i,T_{\mathrm{sim}})\)
    \item \coordbar\coordbar\coordkw{if} import check \(e_i\) passes \coordkw{then}
      \(\mathcal L\gets\mathcal L\cup\{(S_i,A_i)\}\)
    \item \coordbar\coordkw{end if}
    \item \coordkw{end for}
    \item \coordkw{return} \(\mathcal L\)
  \end{enumerate}

  \kern3pt\hrule height .8pt\relax
\end{figure}


\subsection{Evidence Construction}
\label{app:evidence}

\noindent\textbf{Equipment registry.}
The Evidence Agent builds an equipment registry incrementally to
associate similar vessels across occlusion and viewpoint changes.
The registry links observations to apparatus, their owned components,
and consumables, and records contents, states, and demonstrated events.
Consumables are linked to apparatus through use relations.
Robot arms and grippers are recorded as operating context.
Program-assigned identifiers connect each instance to its observations
and annotations.

\noindent\textbf{Frame sampling and association order.}
Base frames are sampled at a fixed temporal interval throughout
each video.
Each frame retains its source video, camera, and decoded timestamp,
and localized observations carry image coordinates.
For multi-camera recordings with a designated primary view,
the agent initializes the registry from primary-camera frames
after an overview pass.
It then scans the primary video over its full duration to establish
the inventory of independent apparatus and consumables.
Observations from the remaining cameras provide complementary views
of these instances and their parts.

\noindent\textbf{Evidence acquisition.}
The agent resolves ambiguous associations by comparing candidate
instances of the same role using reference views and target crops.
It can request an additional crop or a frame at a selected time
to inspect identity, structure, or appearance.
Unresolved associations are retained as explicit hypotheses and
reassessed as new evidence becomes available.

\noindent\textbf{Part-level detail.}
Each instance is annotated with its name, category, owned parts,
demonstrated operations, and functional requirements, including
openings, cavities, detachable parts, joints, valves, measurement,
and support.
The agent uses temporally ordered frames within each clip to link
demonstrated operations to component motion and state changes.
Observed features and requirements cite their supporting frames,
and inferred annotations record the relevant knowledge source
or engineering assumption.
New part-level findings inform the reassessment of pending
association hypotheses.

\noindent\textbf{Laboratory-equipment knowledge.}
\emph{Category rules} describe the characteristic morphology and
functional parts of each apparatus family.
For example, a separatory funnel is characterized by a reservoir
and a lower stopcock.
\emph{Material profiles} represent each part's material as
$m=(p,\mathcal{A})$.
The profile $p$ selects a substrate from a versioned catalog that
records parameter values and their sources.
Density and refractive index use manufacturer references or
documented engineering defaults, depending on the profile.
Shader roughness and transmission weight are rendering controls.
Contact parameters use engineering defaults for dry contact.
The appearance scope $\mathcal{A}$ specifies the tint and finish
of visible regions, including the substrate, coating, ink, and label.
The \emph{transparent-glass policy} assigns a neutral substrate tint
to clear, colorless glass and records colors from the background
or contents as observation context.
During simulation integration, mass is computed from the constructed
geometry and catalog density, as described in Supp.~\ref{app:sim}.

\noindent\textbf{Evidence bundle.}
The registry, reference frames, functional requirements, and resolved
materials form the frozen evidence bundle $\mathcal{D}$.
$\operatorname{CompileTasks}$ compiles this bundle into the
construction tasks $\tau_i$.

\subsection{Asset Construction}
\label{app:asset}

\noindent\textbf{Construction task.}
$\operatorname{CompileTasks}$ compiles the evidence bundle $\mathcal{D}$
into construction tasks $\tau_i$.
Each task specifies the target assets and their owned components,
together with source identifiers, reference frames, functional
requirements, demonstrated interactions, materials, and acceptance criteria.
Density, refractive index, and contact parameters are inherited from
the material profiles.
The agent uses the video evidence to select regional base color and
roughness, and transmission weight for transparent or translucent
glass and plastic surfaces.

\noindent\textbf{Build plan.}
The build plan $\mathcal{B}_i=(\mathcal{N}_i,\mathcal{E}_i)$
defined in Eq.~\eqref{eq:build_plan} specifies named parts $\mathcal{N}_i$
with dimensions, cavities, appearance regions, and material assignments.
The connections $\mathcal{E}_i$ specify joint types, axes, and limits.
Each part is associated with the registry instance or component
it represents.
A component can be realized by several modeling parts, such as the
panels and handles of a balance draft shield.

\noindent\textbf{Plan admission.}
The plan is evaluated against the task for part coverage, material
assignments, cavities, and shared fit constraints.
The resulting feedback identifies the entries to revise.
The agent updates these entries or submits a new draft for structural
revisions, then resubmits the plan.
Construction begins after the plan is admitted.

\noindent\textbf{Construction and sealing.}
The Asset Agent operates Blender through Model Context Protocol
(MCP) tools.
It writes and executes modeling code, queries the scene, assigns
materials to appearance regions, and uses rendered views to refine
the candidate.
The submitted Blender file (\texttt{.blend}), exported meshes
(GLB and OBJ), modeling code, and previews form the source package $S_i$.
Sealing collects these files and records their task, plan, part mappings,
material bindings, and interfaces in a manifest.
In Alg.~\ref{alg:coordinator}, $\mathrm{Build}$ comprises planning,
admission, construction, and sealing, returning
$(\mathcal{B}_i^{(0)},S_i^{(0)})$.
$\mathrm{Revise}$ updates the plan and construction program and
submits the revised, sealed candidate to the Verifier.

\subsection{Independent Verification}
\label{app:verify}

\noindent\textbf{Check targets.}
Let $S_i^{(t)}$ denote the sealed source candidate after $t$ repairs,
with corresponding build plan $\mathcal{B}_i^{(t)}$
and initial submission $S_i^{(0)}$.
The Verifier evaluates the sealed candidate against $\tau_i$,
$\mathcal{B}_i^{(t)}$, and the original video references.
The task and plan define the applicable required targets $\mathcal{K}_i$.
Each target $k=(c,s,r)$ identifies a check $c$, a source item $s$,
and an appearance region $r$ where relevant.
Each receives an outcome
$y_i(k)\in\{\mathrm{Pass},\mathrm{Fail},\mathrm{Unresolved}\}$.
Source acceptance follows Eq.~\eqref{eq:gate}.

\noindent\textbf{Geometric readback.}
A fresh Blender process opens the sealed candidate and measures
part coverage, dimensions, mesh validity, opening and cavity
connectivity, material bindings, and appearance regions.
The Verifier also compares the source geometry with its exported meshes.

\noindent\textbf{Render-vs-video review.}
The Verifier renders views according to the task requirements,
including canonical views, close views along declared openings,
isolated component views, and articulated assembly poses within
the planned joint ranges.
The reviewer compares these renders with the video references along
six appearance dimensions---identity, openings, color layout,
optical surface, surface integrity, and view consistency---and
cites the images supporting each target's outcome.

\noindent\textbf{Evidence gaps.}
When the available evidence leaves a target unresolved, the Verifier
requests additional video crops, renders, alternative backgrounds,
or cross-sections and reassesses the target.
Confirmed failures generate revision requests.
After the evidence-acquisition budget is exhausted, an unresolved
target may also generate a localized revision request.
Its outcome remains $\mathrm{Unresolved}$ until reassessment.

\noindent\textbf{Localized repair.}
The requested edits $F_i$ identify the targets, supporting evidence,
and fields or regions to revise.
The Asset Agent updates the plan and construction program accordingly,
\begin{equation}
  \big(\mathcal{B}_i^{(t+1)},S_i^{(t+1)}\big)
  =
  \mathrm{Revise}\big(\mathcal{B}_i^{(t)},S_i^{(t)},F_i\big),
  \label{eq:repair}
\end{equation}
and submits the sealed successor for verification against all
applicable targets in $\mathcal{K}_i$.
The repair prompt presents the current findings and affected regions,
with earlier records retrieved as needed.
The loop allows at most $T$ source repairs per asset, as set in
Sec.~\ref{sec:setup}, and ends earlier on source acceptance
or when the Verifier returns no requested edits ($F_i=\emptyset$).

\subsection{Verification and Repair Example}
\label{app:verify_example}

\begin{figure}[t]
  \centering
  \definecolor{vfail}{HTML}{873B29}
  \definecolor{vpass}{HTML}{17614D}
  \footnotesize
  \setlength{\tabcolsep}{1.5pt}
  \newcommand{\vw}{\dimexpr(\columnwidth-6pt)/3\relax}
  \begin{tabular}{@{}>{\raggedright\arraybackslash}p{\vw}>{\centering\arraybackslash}p{\vw}>{\centering\arraybackslash}p{\vw}@{}}
    \multicolumn{1}{@{}c}{\textbf{Video reference}} & \textbf{Before repair} & \textbf{After repair} \\[1pt]
    \includegraphics[width=\vw]{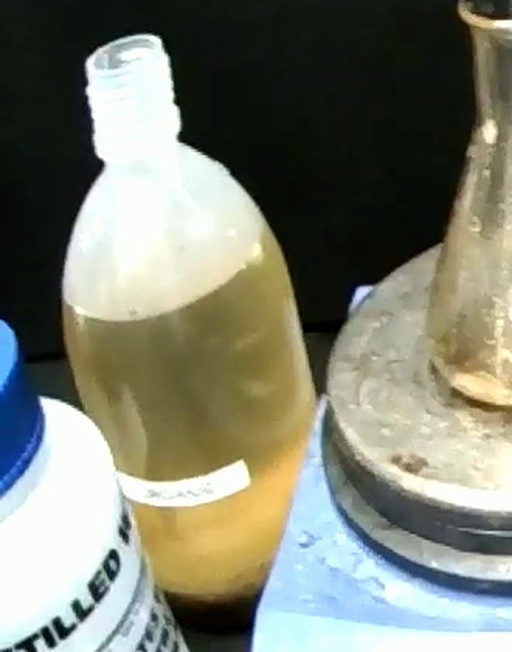} & \includegraphics[width=\vw]{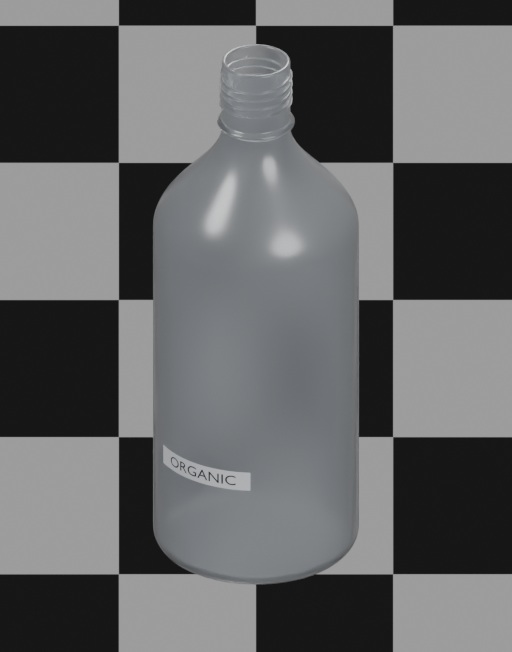} & \includegraphics[width=\vw]{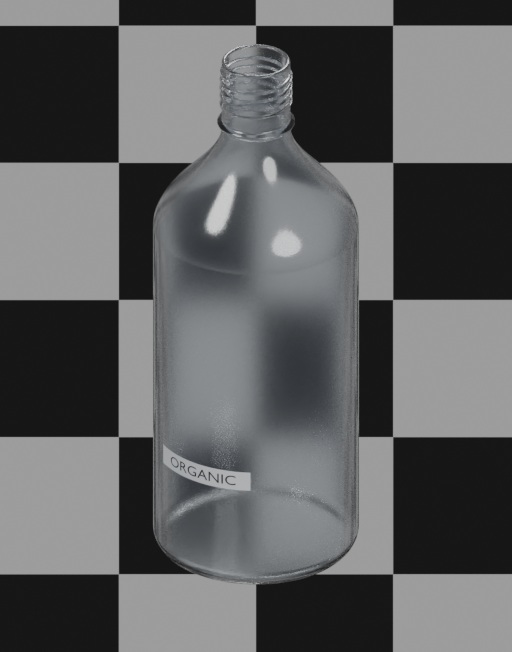} \\[1pt]
    Optical-surface check & \textcolor{vfail}{\textbf{Fail}} & \textcolor{vpass}{\textbf{Pass}} \\
    \arrayrulecolor{black!30}\midrule[0.4pt]\arrayrulecolor{black}
    Wall roughness & 0.22 & 0.12 \\
    Transmission weight & 0.95 & 1.00 \\
  \end{tabular}
  \caption{\textbf{Independent verification and localized repair.} A crop from the operation video shows liquid visible through the wall of a narrow-neck bottle. The Verifier renders the bottle against a checkerboard: before repair, the wall obscures the pattern and the optical-surface check fails. After the Asset Agent lowers the wall roughness and raises its transmission weight, the pattern shows through the body and the check passes.}
  \label{fig:verify}
\end{figure}

Fig.~\ref{fig:verify} illustrates one round of localized repair of a narrow-neck bottle.
Following the Verifier's optical-surface feedback, the Asset Agent adjusts the wall shader, and the revised candidate passes the same check.


\subsection{Simulation Instance Generation}
\label{app:sim}

The Simulation Integration Agent converts each source package $S_i$
into the simulation asset $A_i$ defined in Eq.~\eqref{eq:asset},
using the exported part geometry for $\mathcal{G}_i$.

\noindent\textbf{Mass properties.}
Each volumetric part forms a material partition with the substrate
density specified by its bound profile $p$.
Its mass is computed from the watertight material mesh using
Eq.~\eqref{eq:mass}, and its center of mass and inertia follow
from the same signed tetrahedral integration.
The oriented cavity boundaries subtract the enclosed void,
yielding the mass of the vessel wall.
The mass properties of parts connected by fixed joints are
aggregated using mass-weighted centers and the parallel-axis
theorem to obtain $\mathcal{P}_i$.

\noindent\textbf{Collision.}
For static and kinematic concave bodies, triangle-mesh collision
represents openings and cavities.
Dynamic links use convex collision shapes or collections
of convex pieces.
The agent uses local geometry tools to inspect the meshes
and generate and evaluate candidates for $\mathcal{C}_i$.

\noindent\textbf{Simulator materials.}
For $\mathcal{M}_i$, each part inherits static friction,
dynamic friction, and restitution from its material profile $p$.
The profile's optical parameters and the appearance scope
$\mathcal{A}$ determine the render materials, which are assigned
to the corresponding visual mesh regions.

\noindent\textbf{Joint-motion presets.}
The movable connections in $\mathcal{E}_i$ are compiled into
simulator joints and motion constraints $\mathcal{J}_i$ through
presets selected by connection type.
Each preset specifies the motion model and control behavior,
with joint frames, axes, and limits initialized from
$\mathcal{E}_i$.
Revolute joints (\eg stopcocks and knobs) and prismatic joints
(\eg draft-shield doors) use initial drive gains derived from
the mass and inertia of the moving subtree.
Button return behavior is configured through a return spring
on the prismatic joint.
Screw closures use a native screw constraint that couples
rotation $\theta$ and axial travel $z$ according to
Eq.~\eqref{eq:screw_motion} and is solved together with contacts.
The coupling applies between the seated and disengagement
configurations.
At disengagement, the closure becomes a free body;
it can re-engage during return motion when its relative pose
and velocity satisfy the engagement conditions.

\noindent\textbf{Native execution check.}
Each asset is instantiated in a RoboTwin equipment scene
on SAPIEN.
The execution loads and steps the asset, exercises its prescribed
motions with joint-position readback, renders it, and resets
the scene.
An independent check evaluates material properties, collision
geometry, and joint states from simulator readbacks and compares
simulator renders with the original references.
These results form the import-check report $e_i$
in Alg.~\ref{alg:coordinator}.

\noindent\textbf{Simulation adaptation.}
After the initial import check, the agent performs up to
$T_{\mathrm{sim}}$ adaptation attempts in response to failures
in $e_i$.
Recorded observations, geometry inspections, and diagnostic
executions guide updates to joint drives, return springs,
control modes, screw constraints, collision shapes,
fixed-part overlap treatment, and scene layout.
Ordinary drive adjustments range from $1/16$ to $16$ times
the initial mass-derived stiffness, damping, and force limit.
These updates act on the simulation representation of $A_i$.
Adaptation proceeds from the current candidate while maintaining
the best retained result.
The retained result is updated when a candidate preserves
previously passed checks and achieves one of the following
under the same execution protocol: a later execution stage,
a reduced failed-check set at the same stage, or improved
failure metrics with the stage and failed-check set unchanged.
The loop ends when the import check passes, the agent stops,
or the adaptation budget is exhausted, and returns the retained
result with its import-check report.


\section{Dataset and Implementation Details}
\label{app:data}

\begin{table}[t]
  \centering
  \small
  \caption{\textbf{Reference annotations.} Equipment instances, required parts, and required part--action pairs of each scene.}
  \label{tab:supp_data}
  \setlength{\tabcolsep}{4pt}
  \begin{tabular}{llccc}
    \toprule
    Scene & Video & \#Inst. & \#Parts & \#Actions \\
    \midrule
    1 (Ego)   & 57 clips, 228\,s  & 18 & 54 & 41 \\
    2 (Robot) & 3 cameras, 130\,s & 11 & 29 & 7 \\
    \bottomrule
  \end{tabular}
\end{table}

\noindent\textbf{Scenes.}
Scene~1 consists of 57 egocentric clips, each 4\,s long at 4 frames per second, from a public collection of laboratory-action videos~\cite{ibm2024labactions}; together they show a workflow of weighing, reagent addition, stirring, phase separation, layer collection, and pH measurement.
Its 18 reference instances are an analytical balance with draft-shield doors, a separatory funnel with a stopcock and its ring stand, a hotplate stirrer with two knobs, a stir bar, a pH meter, a graduated cylinder, an Erlenmeyer flask, a sample beaker, a rinse beaker, a ribbed reagent bottle, bottles for the aqueous and organic phases, a powder jar, a wash bottle, a spoon, weighing paper, and a paper packet.
Scene~2 is a robot weighing-and-transfer setup recorded for 130\,s by three synchronized head and wrist cameras.
Its 11 reference instances are a balance, a circular platform, a transfer funnel, a long-stem funnel, a funnel stand, a graduated cylinder, an Erlenmeyer flask, a small beaker, a white bowl, a scoop, and weighing paper.
People, robots, benches, and background are excluded from both inventories.
Tab.~\ref{tab:supp_data} summarizes the shared reference annotation of each scene, as defined in Supp.~\ref{app:protocol}.

\noindent\textbf{Baselines.}
Each Codex baseline is asked to deliver, for every piece of equipment, an editable Blender model, visual meshes, collision geometry, mass properties, and joints loadable in SAPIEN/RoboTwin.
It works from the same source videos and Blender MCP server as \method and decides by itself how to inspect the videos and how to check its output.
Real2Gym~\cite{ren2026real2gym} reconstructs editable simulation scenes from human and robot demonstrations; applied to the same videos, it delivers MuJoCo models, which are executed natively in MuJoCo, as described in Supp.~\ref{app:protocol}.
SimFoundry~\cite{ranawaka2026simfoundry} is a modular video-to-simulation pipeline that extracts objects from the video, generates a 3D mesh for each object, and compiles the meshes into a physical scene.
It receives the complete workflow video from the primary camera view of each scene, and its delivered objects are loaded into the same RoboTwin scene as the assets of \method and the Codex baselines.

\noindent\textbf{Implementation.}
All four modules of \method use GPT-6-Astra with
module-specific reasoning effort.
The Evidence Agent uses \emph{low} effort for scene overview
and frame scanning, and \emph{medium} effort for registry
initialization, observation association, and part-level
detail extraction.
The Asset Agent and Simulation Integration Agent use
\emph{xhigh} effort, while the Verifier uses \emph{high} effort.
These settings are used in both scenes.
Both Codex baselines also use \emph{xhigh} reasoning effort.

Assets are built in Blender~5.2 and validated in RoboTwin on SAPIEN.
Each asset receives up to $T{=}5$ source repairs in the verification loop of Supp.~\ref{app:verify} and up to $T_{\mathrm{sim}}{=}5$ simulation adaptation attempts in the integration stage of Supp.~\ref{app:sim}.
All experiments run on a single NVIDIA RTX~4090\,D GPU.

\noindent\textbf{Ablation variants.}
\emph{images-input} gives the agents 25 selected frames from Scene~1 and three frames from each of Scene~2's three cameras in place of the videos, with all downstream components unchanged.
\emph{w/o feedback} keeps the Asset Agent's initial submission as the source asset, $S_i=S_i^{(0)}$, without Verifier-guided correction or repair.
\emph{w/o adaptation} and \emph{w/o joint presets} reuse the source assets $S_i$ of the full framework.
\emph{w/o adaptation} removes the model's initial configuration adjustments and the feedback-driven adaptation loop ($T_{\mathrm{sim}}{=}0$).
\emph{w/o joint presets} replaces the joint-motion presets described in Supp.~\ref{app:sim} with native joint configurations that the Simulation Integration Agent authors for each movable connection in $\mathcal{E}_i$.

\section{Evaluation Protocol}
\label{app:protocol}

Within each scene, all methods use the same reference annotations, action definitions, rendering protocol, and scoring rules.
We report the four dimensions separately.

\noindent\textbf{Shared reference.}
For each scene, a reference annotation is built from the source videos and shared by all methods, as summarized in Tab.~\ref{tab:supp_data}.
It lists the equipment instances, their visible features, the required parts, and the required part--action pairs with completion criteria such as reaching the end of travel, closing, separating, re-seating, or springing back.
Required actions include both operations demonstrated in the videos and operations implied by the visible structure; each records its basis.

\noindent\textbf{A: Instance completeness.}
Reference instances are matched one-to-one to generated physical instances, where one delivery may contain several instances and several deliveries may form one instance.
With $V$ reference instances, $G$ generated instances, and $M$ matches, instance completeness is measured by the intersection-over-union of the two instance sets:
\begin{equation}
  A = 100\,\frac{M}{V+G-M}.
  \label{eq:eval_a}
\end{equation}
This score accounts for missing, extra, and duplicated instances.

\noindent\textbf{B: Visual fidelity.}
Each delivered asset is rendered in Blender Cycles from four orthographic views under a common lighting setup scaled to its bounding box, with close-ups of relevant parts.
Comparing the renders with the source frames, the judge assigns three integer ratings in $[1,10]$ against fixed anchors: \emph{appearance} (silhouette, proportions, and printed markings), \emph{material} (color, transparency, gloss, and frosting), and \emph{visible structure} (3D layout of necks, openings, valves, and knobs).
A missing instance receives 0 for all three ratings; a subscore whose defining visible features are wholly inconsistent also receives 0.
With $\bar r$ denoting the mean rating over applicable reference instances,
$B = 3.5\,\bar r_{\mathrm{app}} + 3.5\,\bar r_{\mathrm{mat}} + 3\,\bar r_{\mathrm{str}}$.

\noindent\textbf{C: Parts and articulation.}
Each required part receives a presence credit $p_j\in\{0,1\}$, and each required action receives a completion credit $a_l\in\{0,0.5,1\}$: 1 if the part moves in the required way and completes the operation, 0.5 if the motion type is correct but the operation is only partially completed because of insufficient travel or jamming, and 0 otherwise.
With $N_p$ required parts and $N_a$ required actions,
\begin{equation}
  C =
  \underbrace{\frac{50}{N_p}\sum_{j=1}^{N_p} p_j}_{C_{\mathrm{p}}}
  +
  \underbrace{\frac{50}{N_a}\sum_{l=1}^{N_a} a_l}_{C_{\mathrm{m}}}.
  \label{eq:eval_c}
\end{equation}
For \method, the Codex baselines, and SimFoundry, actions are executed in a native RoboTwin scene (SAPIEN PhysX, $\Delta t = 1/240$\,s) with finite-force joint or relative-pose drives calibrated to the joint frames of each delivery.
Each annotated operation receives one action credit based on its complete motion sequence, judged from the continuous simulation recording.
In Scene~1, cap removal and replacement are scored separately, while pressing a meter button and observing its spring return constitute one action.
When incomplete travel may arise from the test drive, we evaluate two predefined higher-gain controller profiles under the same force limits.
Real2Gym delivers MuJoCo models, which are executed natively in MuJoCo~\cite{todorov2012mujoco} with their own collision geometry, mass, and inertia, under the same action definitions and scoring rules.

\noindent\textbf{D: Valve-controlled drainage.}
For every separatory funnel, the largest connected cavity of the delivered geometry is filled to half its height with liquid simulated by MLS-MPM on the GPU; a funnel without a connected cavity cannot be filled and fails.
The valve is held closed for 2\,s, opened over 1\,s, and held open for 4\,s.
A funnel passes if no sustained outflow occurs while the valve is closed and liquid visibly drains through the lower outlet once it opens; leakage through walls fails.
Liquid--solid interaction is computed against the delivered meshes using swept collision detection.
$D$ is the percentage of passing funnels; Scene~2 contains no separatory funnel, so $D$ is reported for Scene~1 only.

\noindent\textbf{Judging.}
Visual ratings and action credits are assigned by GPT-6-Astra with \emph{max} reasoning effort using source frames, standardized asset renders, and continuous simulation recordings.
Each judgment is linked to its visual evidence, and the evaluation scripts aggregate the ratings and credits into scene-level scores.
The prompts for the four dimensions are given in Supp.~\ref{app:prompts}.

\begin{figure*}[tp]
  \centering
  \colorlet{qband}{cvprblue}
  \definecolor{qours}{RGB}{214,96,0}
  \setlength{\tabcolsep}{0.6pt}
  \setlength{\fboxsep}{0pt}
  \renewcommand{\arraystretch}{0}
  \newcommand{\qw}{0.1\textwidth}
  \newcommand{\qh}{0.084\textwidth}
  \newcommand{\qcell}[1]{{\setlength{\fboxrule}{0.3pt}\fcolorbox{gray!45}{white}{#1}}}
  \newcommand{\qours}[1]{{\setlength{\fboxrule}{0.6pt}\fcolorbox{qours}{white}{#1}}}
  \newcommand{\qmiss}{{\setlength{\fboxrule}{0.3pt}\fcolorbox{gray!45}{gray!10}{\parbox[b][\dimexpr\qh-0.504pt\relax][c]{\dimexpr\qw-0.6pt\relax}{\centering\scriptsize\color{gray}not\\recovered}}}}
  \newcommand{\qrow}[2][\qh]{\rotatebox{90}{\parbox{#1}{\centering\scriptsize #2}}}
  \newcommand{\qhead}[1]{\parbox[b]{\qw}{\centering\scriptsize #1}}
  \newcommand{\qscene}[3]{\multicolumn{#1}{@{}c@{}}{\colorbox{qband}{\makebox[#2]{\rule[-2.3pt]{0pt}{9.6pt}\small\color{white}\textbf{#3}}}}}
  \begin{tabular}{@{}c@{\hspace{2pt}}cccccccc@{}}
    & \qscene{8}{\dimexpr0.8\textwidth+8.4pt\relax}{Scene 1 (Ego)} \\[2.5pt]
    & \qhead{pH meter} & \qhead{Graduated\\cylinder} & \qhead{Erlenmeyer\\flask} & \qhead{Sample\\beaker} & \qhead{Aqueous-phase\\bottle} & \qhead{Organic-phase\\bottle} & \qhead{Powder\\jar} & \qhead{Spoon} \\[2pt]
    \qrow[\qw]{Real world} & \includegraphics[width=\qw]{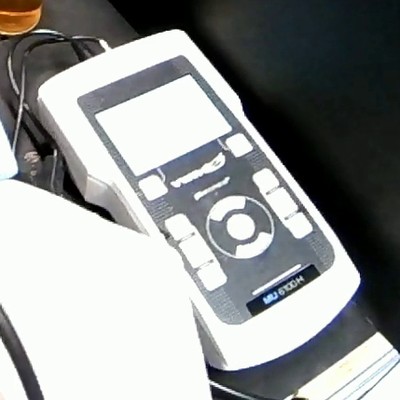} & \includegraphics[width=\qw]{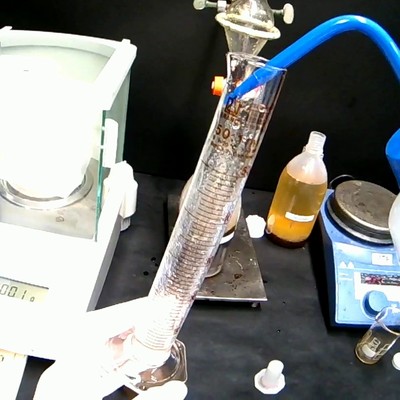} & \includegraphics[width=\qw]{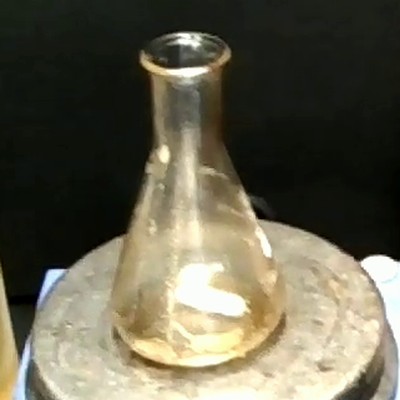} & \includegraphics[width=\qw]{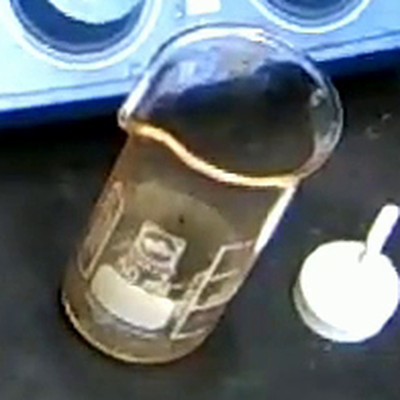} & \includegraphics[width=\qw]{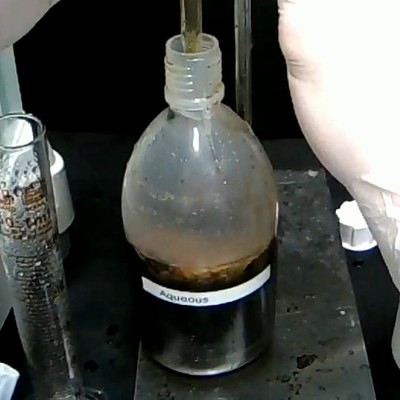} & \includegraphics[width=\qw]{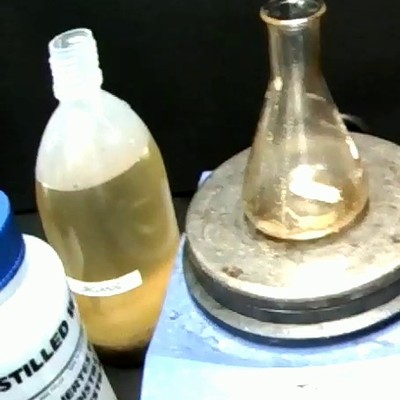} & \includegraphics[width=\qw]{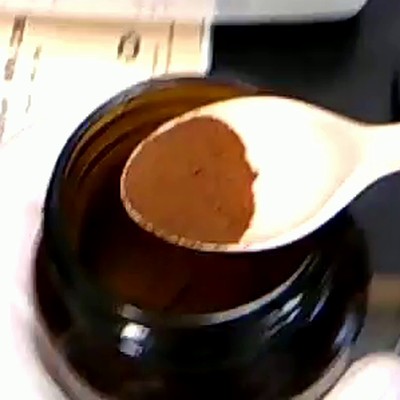} & \includegraphics[width=\qw]{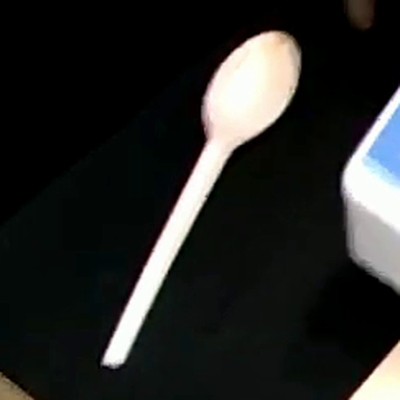} \\[1.2pt]
    \qrow{Real2Gym} & \qcell{\includegraphics[width=\dimexpr\qw-0.6pt\relax,trim=0 32 0 32,clip]{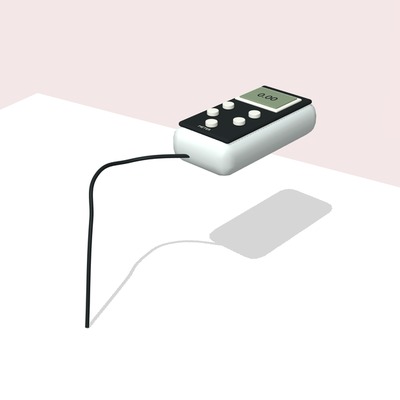}} & \qcell{\includegraphics[width=\dimexpr\qw-0.6pt\relax,trim=0 32 0 32,clip]{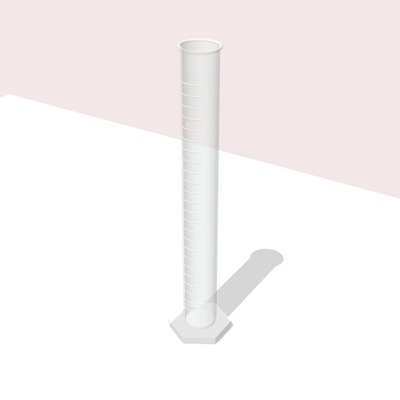}} & \qcell{\includegraphics[width=\dimexpr\qw-0.6pt\relax,trim=0 32 0 32,clip]{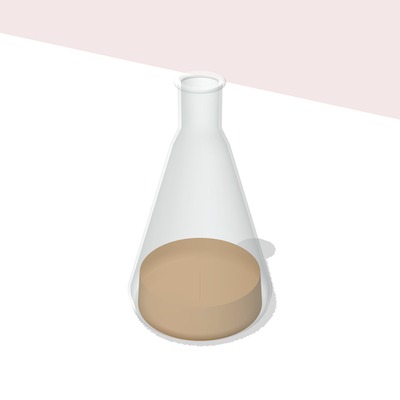}} & \qcell{\includegraphics[width=\dimexpr\qw-0.6pt\relax,trim=0 32 0 32,clip]{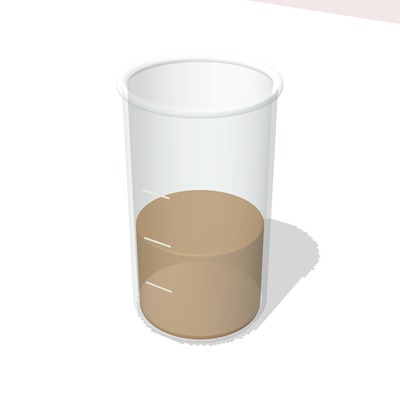}} & \qcell{\includegraphics[width=\dimexpr\qw-0.6pt\relax,trim=0 32 0 32,clip]{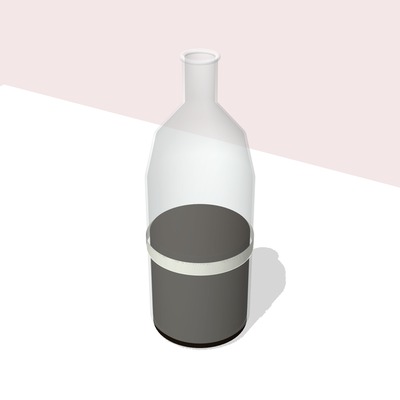}} & \qcell{\includegraphics[width=\dimexpr\qw-0.6pt\relax,trim=0 32 0 32,clip]{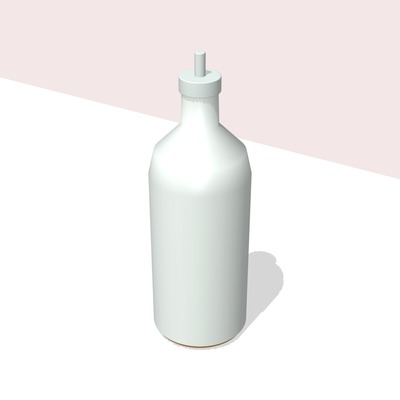}} & \qcell{\includegraphics[width=\dimexpr\qw-0.6pt\relax,trim=0 32 0 32,clip]{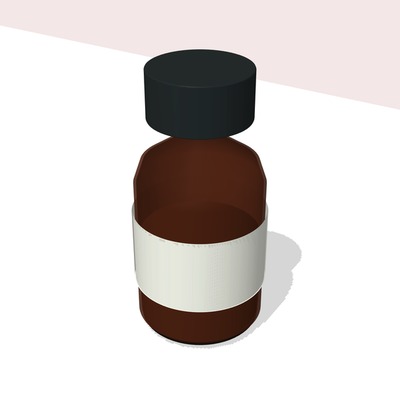}} & \qmiss \\[1.2pt]
    \qrow{\makebox[\linewidth][c]{SimFoundry}} & \qmiss & \qcell{\includegraphics[width=\dimexpr\qw-0.6pt\relax,trim=0 32 0 32,clip]{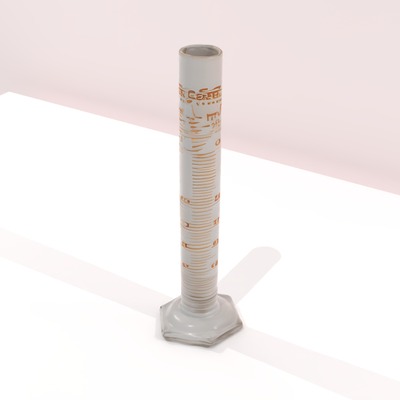}} & \qcell{\includegraphics[width=\dimexpr\qw-0.6pt\relax,trim=0 32 0 32,clip]{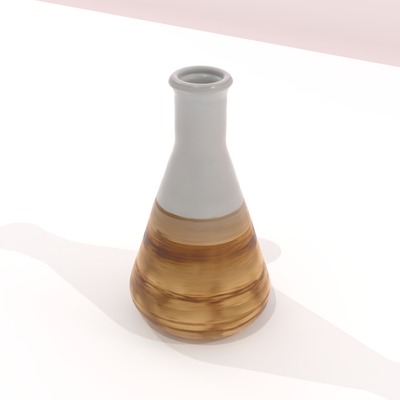}} & \qcell{\includegraphics[width=\dimexpr\qw-0.6pt\relax,trim=0 32 0 32,clip]{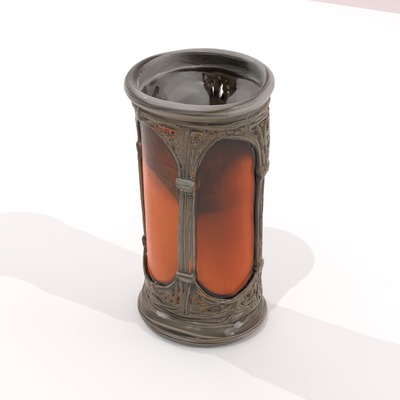}} & \qcell{\includegraphics[width=\dimexpr\qw-0.6pt\relax,trim=0 32 0 32,clip]{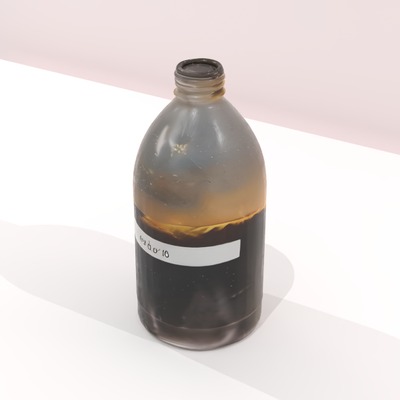}} & \qcell{\includegraphics[width=\dimexpr\qw-0.6pt\relax,trim=0 32 0 32,clip]{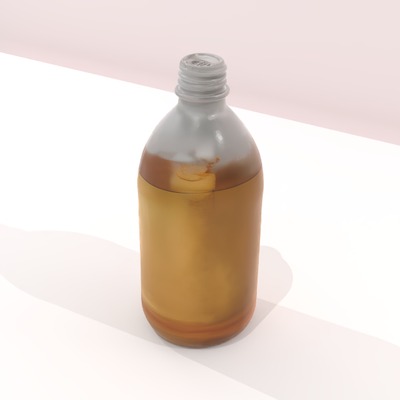}} & \qmiss & \qmiss \\[1.2pt]
    \qrow{Codex +\\\makebox[\linewidth][c]{GPT-6-Astra}} & \qcell{\includegraphics[width=\dimexpr\qw-0.6pt\relax,trim=0 32 0 32,clip]{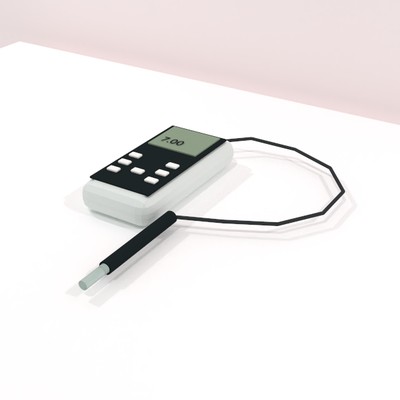}} & \qcell{\includegraphics[width=\dimexpr\qw-0.6pt\relax,trim=0 32 0 32,clip]{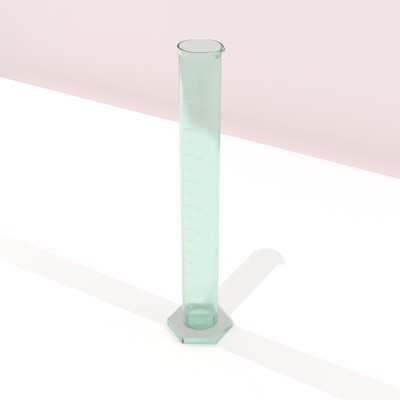}} & \qcell{\includegraphics[width=\dimexpr\qw-0.6pt\relax,trim=0 32 0 32,clip]{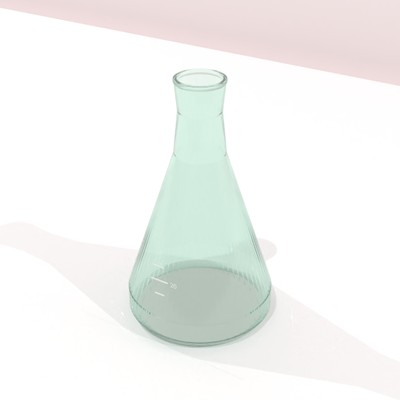}} & \qcell{\includegraphics[width=\dimexpr\qw-0.6pt\relax,trim=0 32 0 32,clip]{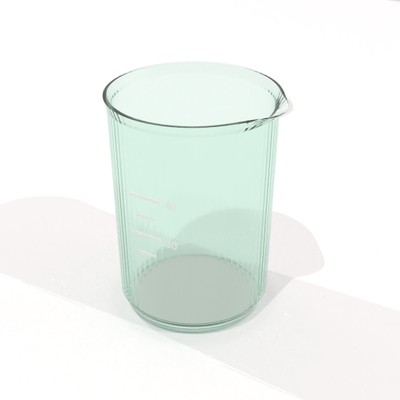}} & \qcell{\includegraphics[width=\dimexpr\qw-0.6pt\relax,trim=0 32 0 32,clip]{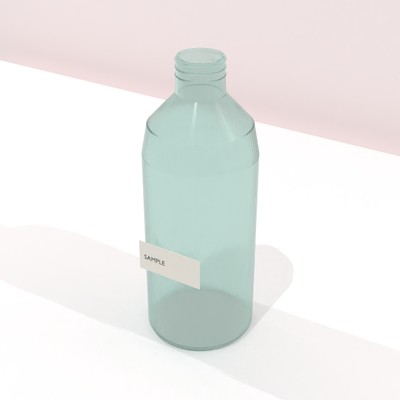}} & \qcell{\includegraphics[width=\dimexpr\qw-0.6pt\relax,trim=0 32 0 32,clip]{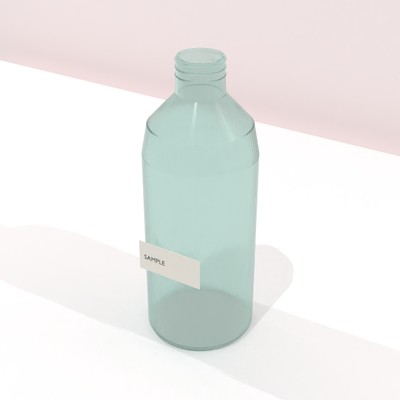}} & \qcell{\includegraphics[width=\dimexpr\qw-0.6pt\relax,trim=0 32 0 32,clip]{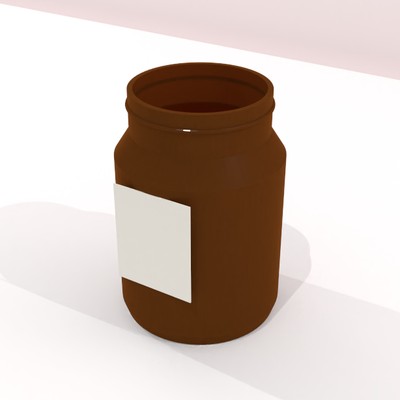}} & \qcell{\includegraphics[width=\dimexpr\qw-0.6pt\relax,trim=0 32 0 32,clip]{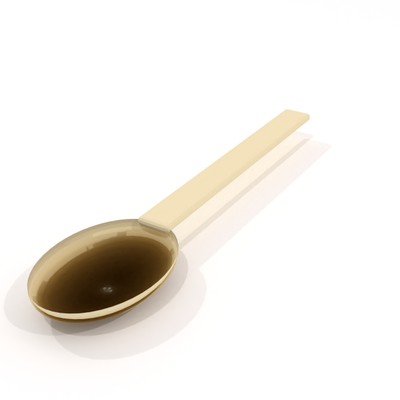}} \\[1.2pt]
    \qrow{Codex +\\GPT-6-Sol} & \qcell{\includegraphics[width=\dimexpr\qw-0.6pt\relax,trim=0 32 0 32,clip]{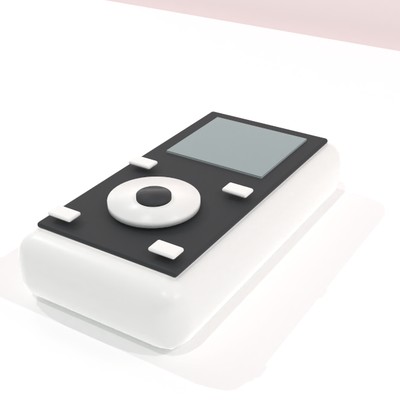}} & \qcell{\includegraphics[width=\dimexpr\qw-0.6pt\relax,trim=0 32 0 32,clip]{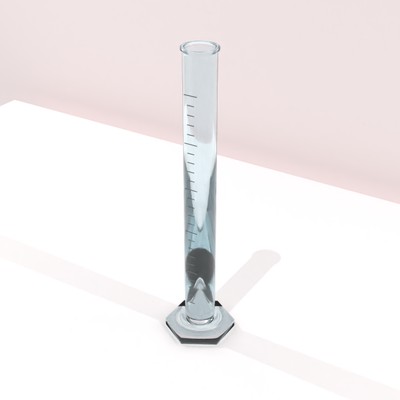}} & \qcell{\includegraphics[width=\dimexpr\qw-0.6pt\relax,trim=0 32 0 32,clip]{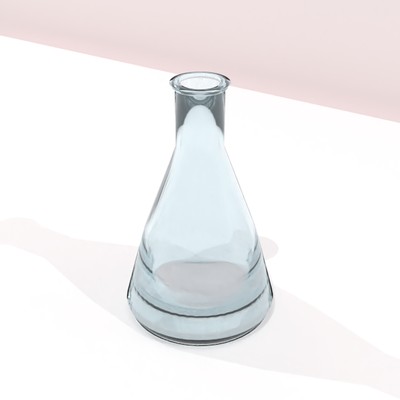}} & \qcell{\includegraphics[width=\dimexpr\qw-0.6pt\relax,trim=0 32 0 32,clip]{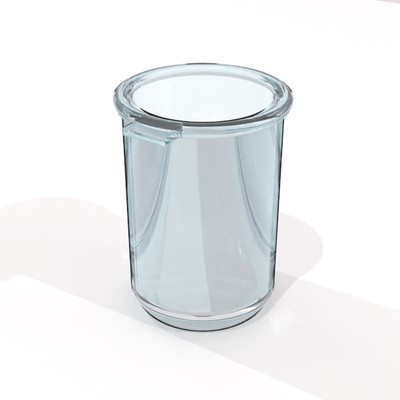}} & \qcell{\includegraphics[width=\dimexpr\qw-0.6pt\relax,trim=0 32 0 32,clip]{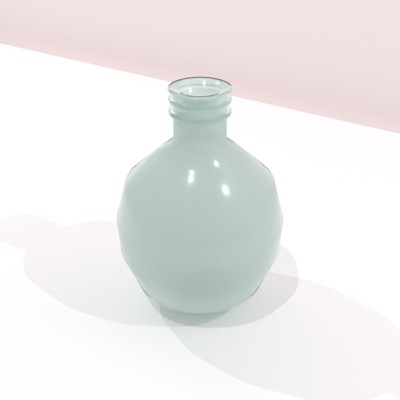}} & \qcell{\includegraphics[width=\dimexpr\qw-0.6pt\relax,trim=0 32 0 32,clip]{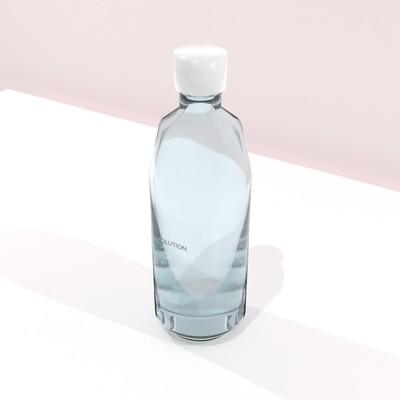}} & \qcell{\includegraphics[width=\dimexpr\qw-0.6pt\relax,trim=0 32 0 32,clip]{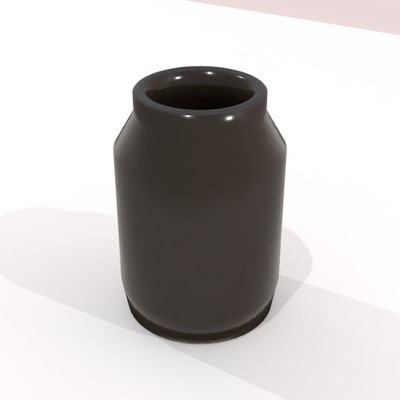}} & \qcell{\includegraphics[width=\dimexpr\qw-0.6pt\relax,trim=0 32 0 32,clip]{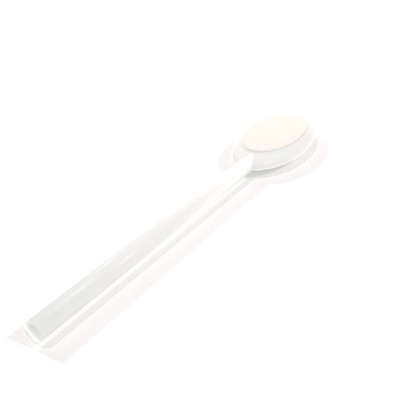}} \\[2.5pt]
    \qrow{\color{qours}\textbf{\method}\\\color{qours}(Ours)} & \qours{\includegraphics[width=\dimexpr\qw-1.2pt\relax,trim=0 32 0 32,clip]{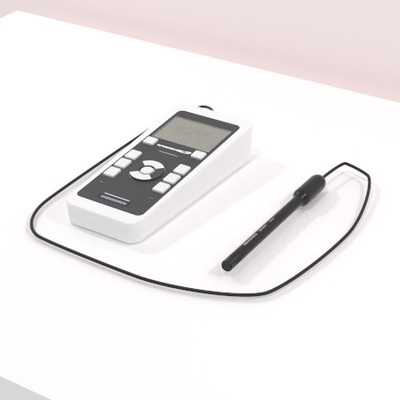}} & \qours{\includegraphics[width=\dimexpr\qw-1.2pt\relax,trim=0 32 0 32,clip]{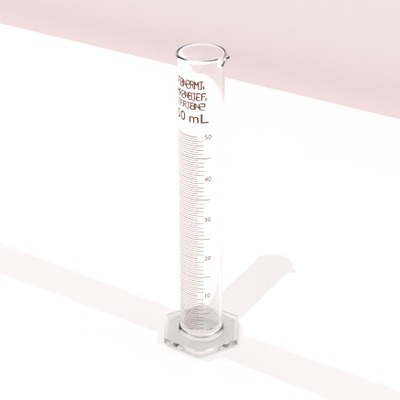}} & \qours{\includegraphics[width=\dimexpr\qw-1.2pt\relax,trim=0 32 0 32,clip]{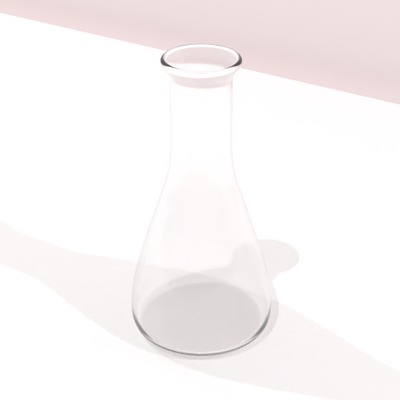}} & \qours{\includegraphics[width=\dimexpr\qw-1.2pt\relax,trim=0 32 0 32,clip]{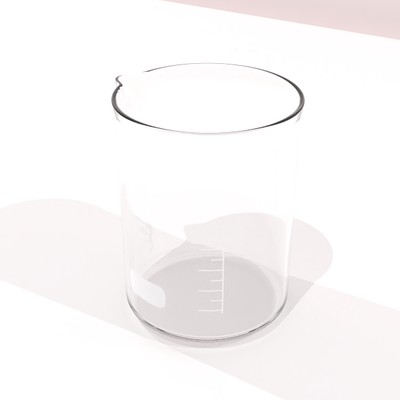}} & \qours{\includegraphics[width=\dimexpr\qw-1.2pt\relax,trim=0 32 0 32,clip]{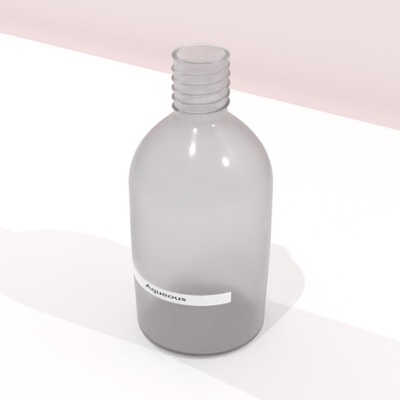}} & \qours{\includegraphics[width=\dimexpr\qw-1.2pt\relax,trim=0 32 0 32,clip]{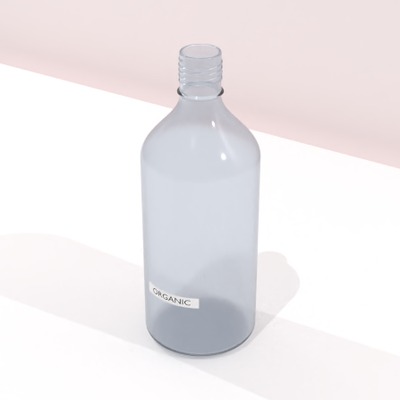}} & \qours{\includegraphics[width=\dimexpr\qw-1.2pt\relax,trim=0 32 0 32,clip]{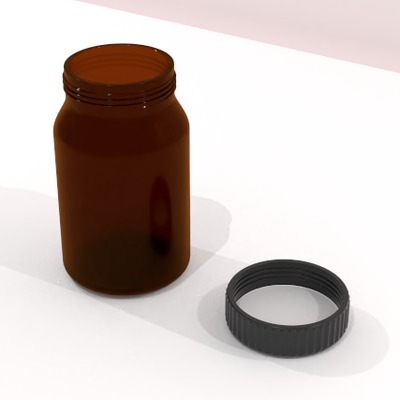}} & \qours{\includegraphics[width=\dimexpr\qw-1.2pt\relax,trim=0 32 0 32,clip]{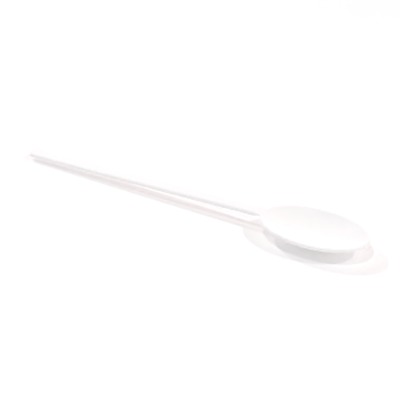}} \\[5pt]
    & \qscene{8}{\dimexpr0.8\textwidth+8.4pt\relax}{Scene 2 (Robot)} \\[2.5pt]
    & \qhead{Long-stem\\funnel} & \qhead{Graduated\\cylinder} & \qhead{Erlenmeyer\\flask} & \qhead{Beaker} & \qhead{Circular\\platform} & \qhead{Bowl} & \qhead{Scoop} & \qhead{Weighing\\paper} \\[2pt]
    \qrow[\qw]{Real world} & \includegraphics[width=\qw]{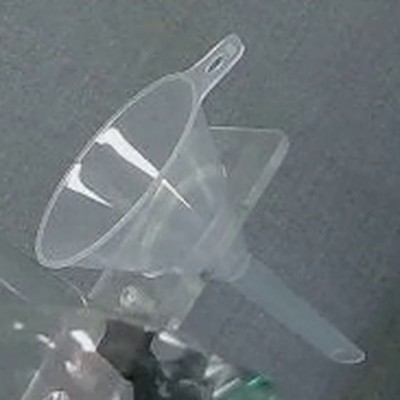} & \includegraphics[width=\qw]{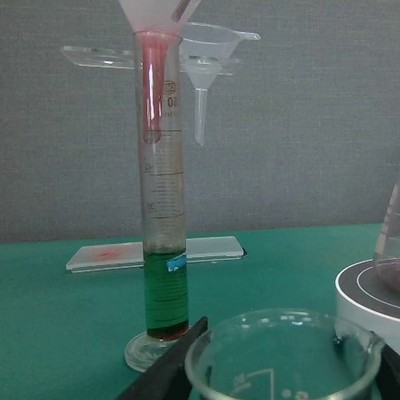} & \includegraphics[width=\qw]{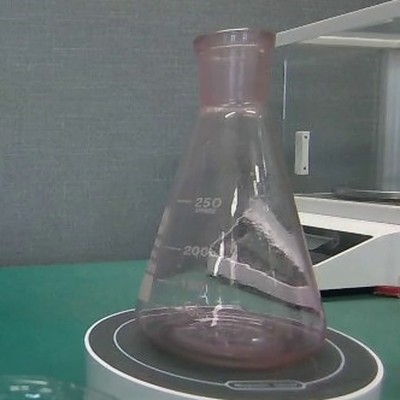} & \includegraphics[width=\qw]{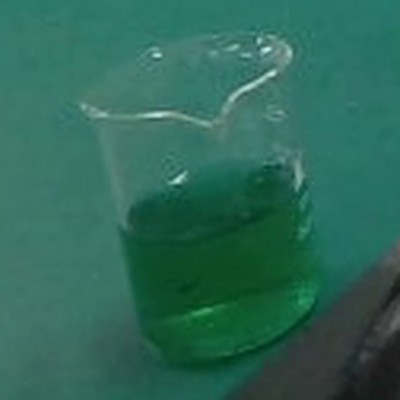} & \includegraphics[width=\qw]{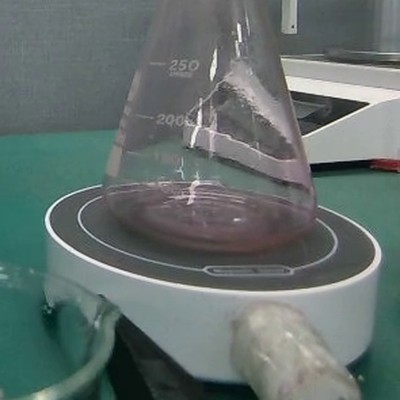} & \includegraphics[width=\qw]{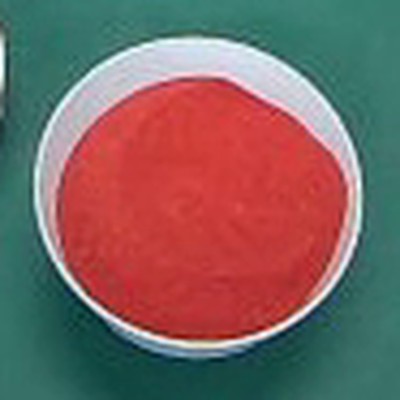} & \includegraphics[width=\qw]{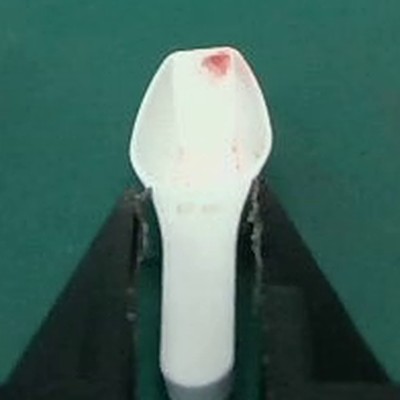} & \includegraphics[width=\qw]{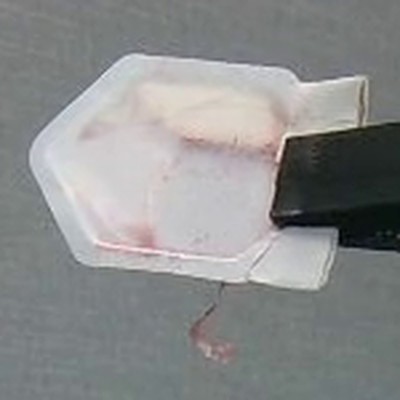} \\[1.2pt]
    \qrow{Real2Gym} & \qcell{\includegraphics[width=\dimexpr\qw-0.6pt\relax,trim=0 32 0 32,clip]{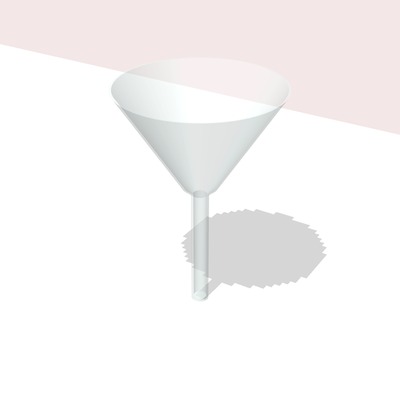}} & \qcell{\includegraphics[width=\dimexpr\qw-0.6pt\relax,trim=0 32 0 32,clip]{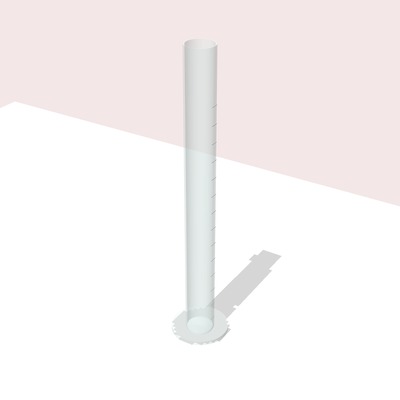}} & \qcell{\includegraphics[width=\dimexpr\qw-0.6pt\relax,trim=0 32 0 32,clip]{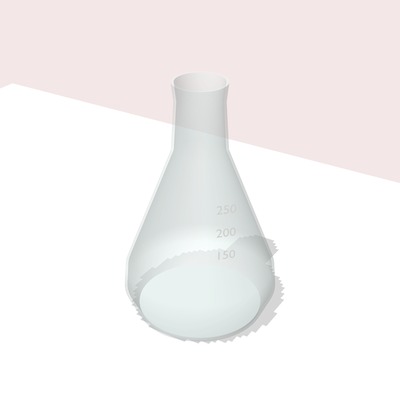}} & \qcell{\includegraphics[width=\dimexpr\qw-0.6pt\relax,trim=0 32 0 32,clip]{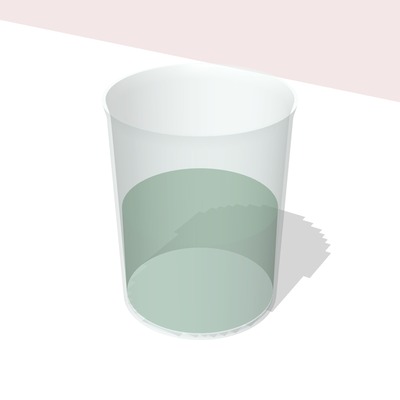}} & \qcell{\includegraphics[width=\dimexpr\qw-0.6pt\relax,trim=0 32 0 32,clip]{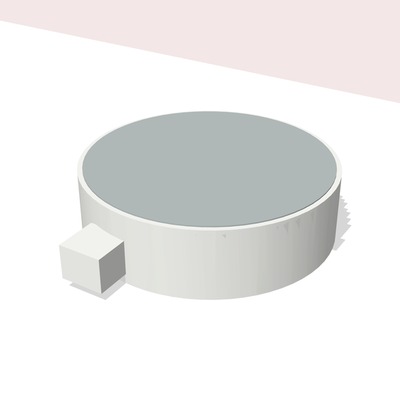}} & \qcell{\includegraphics[width=\dimexpr\qw-0.6pt\relax,trim=0 32 0 32,clip]{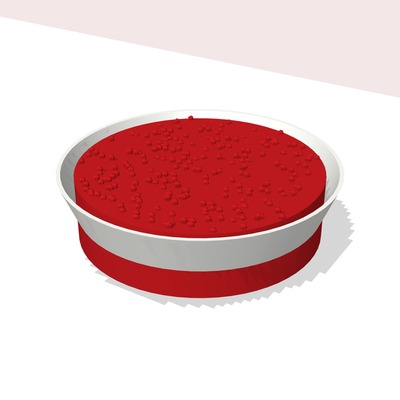}} & \qcell{\includegraphics[width=\dimexpr\qw-0.6pt\relax,trim=0 32 0 32,clip]{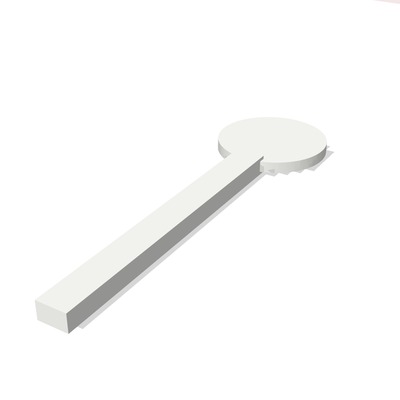}} & \qmiss \\[1.2pt]
    \qrow{\makebox[\linewidth][c]{SimFoundry}} & \qmiss & \qcell{\includegraphics[width=\dimexpr\qw-0.6pt\relax,trim=0 32 0 32,clip]{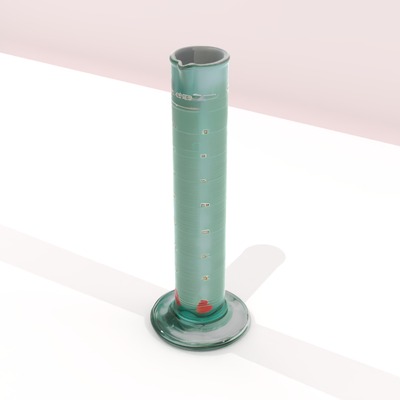}} & \qcell{\includegraphics[width=\dimexpr\qw-0.6pt\relax,trim=0 32 0 32,clip]{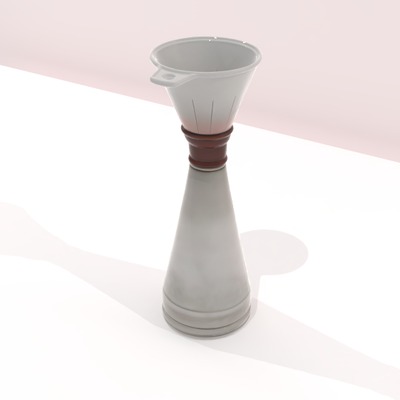}} & \qcell{\includegraphics[width=\dimexpr\qw-0.6pt\relax,trim=0 32 0 32,clip]{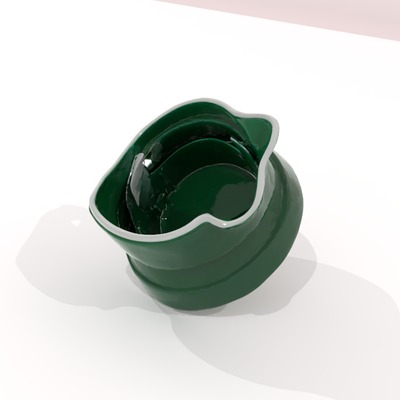}} & \qcell{\includegraphics[width=\dimexpr\qw-0.6pt\relax,trim=0 32 0 32,clip]{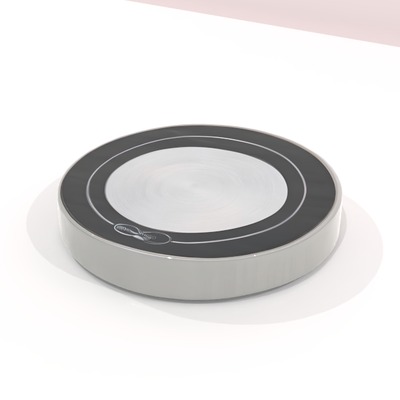}} & \qcell{\includegraphics[width=\dimexpr\qw-0.6pt\relax,trim=0 32 0 32,clip]{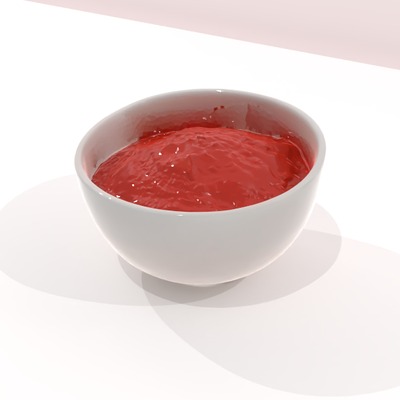}} & \qcell{\includegraphics[width=\dimexpr\qw-0.6pt\relax,trim=0 32 0 32,clip]{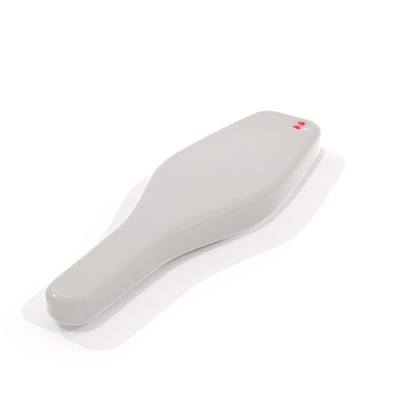}} & \qcell{\includegraphics[width=\dimexpr\qw-0.6pt\relax,trim=0 32 0 32,clip]{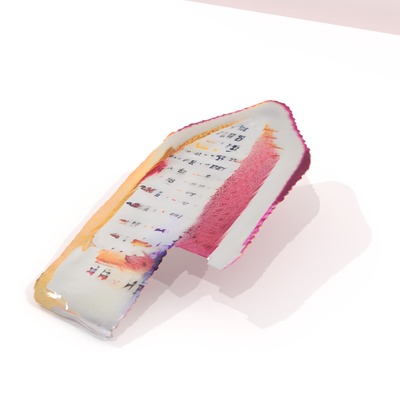}} \\[1.2pt]
    \qrow{Codex +\\\makebox[\linewidth][c]{GPT-6-Astra}} & \qcell{\includegraphics[width=\dimexpr\qw-0.6pt\relax,trim=0 32 0 32,clip]{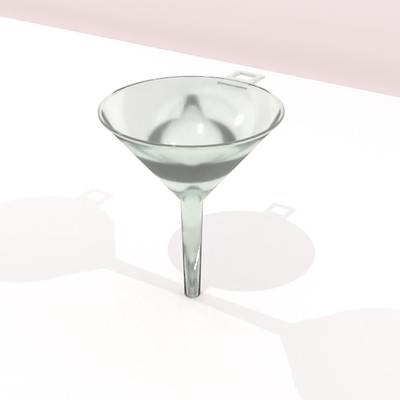}} & \qcell{\includegraphics[width=\dimexpr\qw-0.6pt\relax,trim=0 32 0 32,clip]{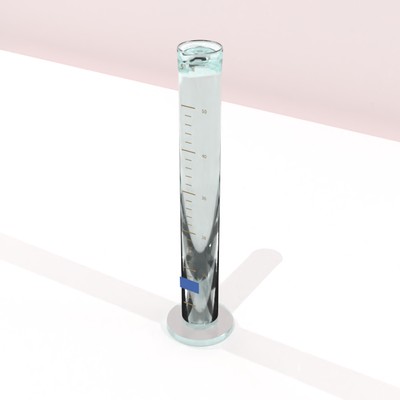}} & \qcell{\includegraphics[width=\dimexpr\qw-0.6pt\relax,trim=0 32 0 32,clip]{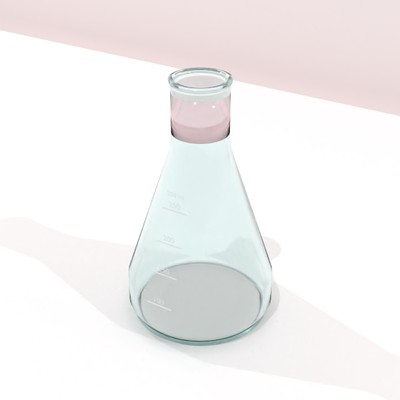}} & \qcell{\includegraphics[width=\dimexpr\qw-0.6pt\relax,trim=0 32 0 32,clip]{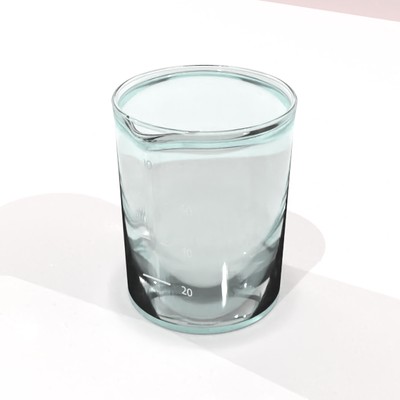}} & \qcell{\includegraphics[width=\dimexpr\qw-0.6pt\relax,trim=0 32 0 32,clip]{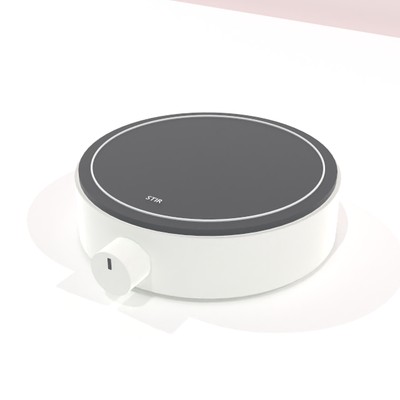}} & \qcell{\includegraphics[width=\dimexpr\qw-0.6pt\relax,trim=0 32 0 32,clip]{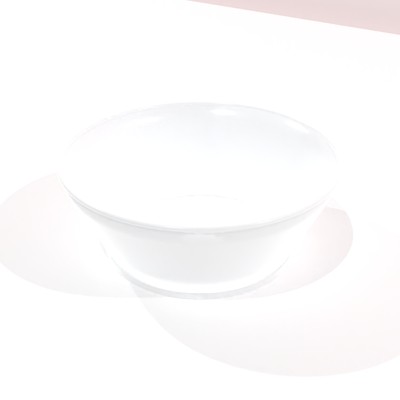}} & \qcell{\includegraphics[width=\dimexpr\qw-0.6pt\relax,trim=0 32 0 32,clip]{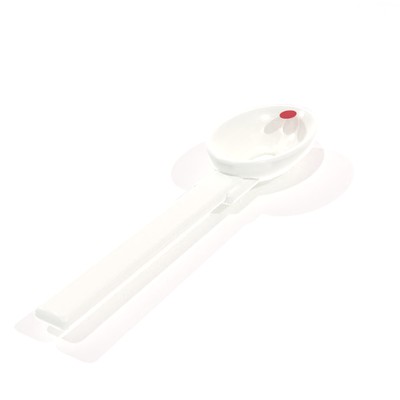}} & \qcell{\includegraphics[width=\dimexpr\qw-0.6pt\relax,trim=0 32 0 32,clip]{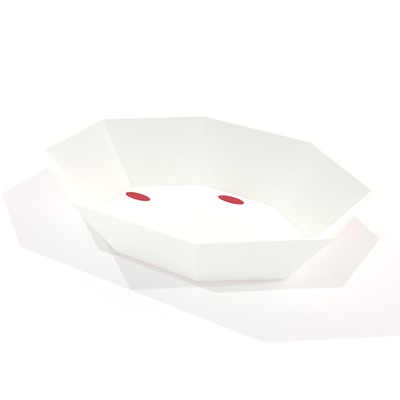}} \\[1.2pt]
    \qrow{Codex +\\GPT-6-Sol} & \qcell{\includegraphics[width=\dimexpr\qw-0.6pt\relax,trim=0 32 0 32,clip]{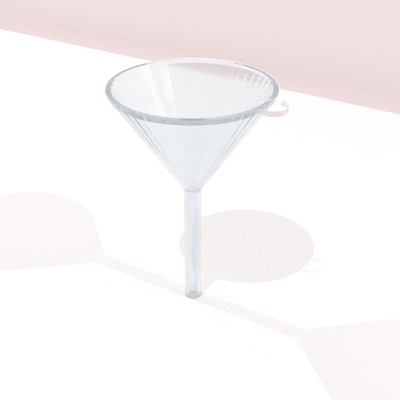}} & \qcell{\includegraphics[width=\dimexpr\qw-0.6pt\relax,trim=0 32 0 32,clip]{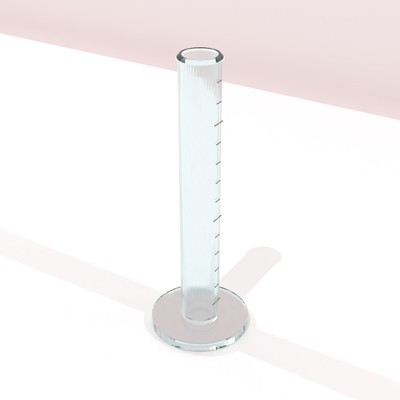}} & \qcell{\includegraphics[width=\dimexpr\qw-0.6pt\relax,trim=0 32 0 32,clip]{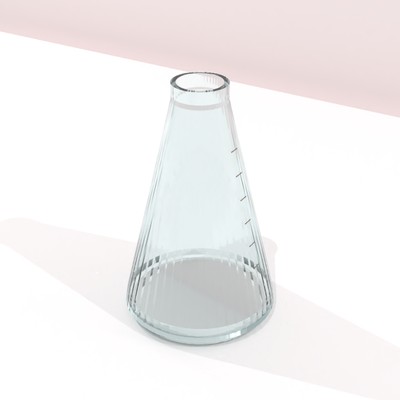}} & \qcell{\includegraphics[width=\dimexpr\qw-0.6pt\relax,trim=0 32 0 32,clip]{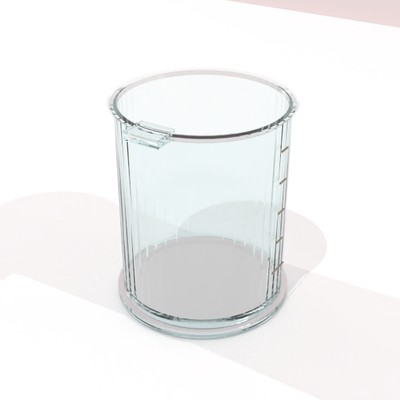}} & \qcell{\includegraphics[width=\dimexpr\qw-0.6pt\relax,trim=0 32 0 32,clip]{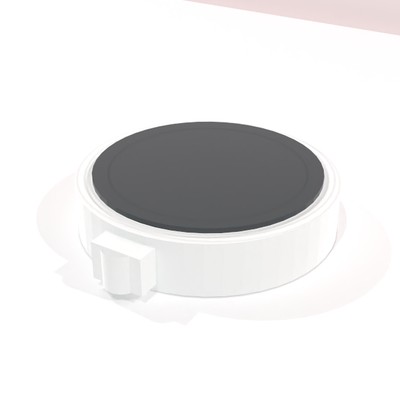}} & \qcell{\includegraphics[width=\dimexpr\qw-0.6pt\relax,trim=0 32 0 32,clip]{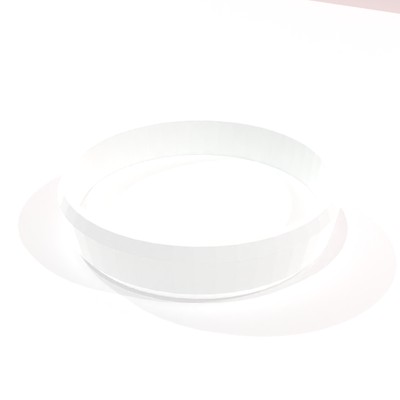}} & \qcell{\includegraphics[width=\dimexpr\qw-0.6pt\relax,trim=0 32 0 32,clip]{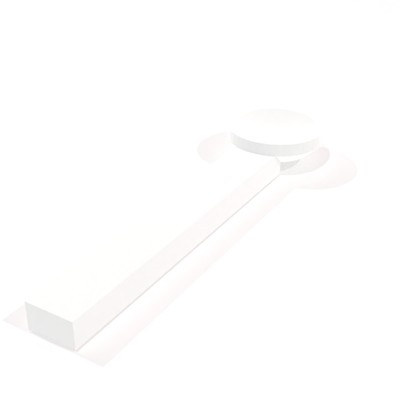}} & \qcell{\includegraphics[width=\dimexpr\qw-0.6pt\relax,trim=0 32 0 32,clip]{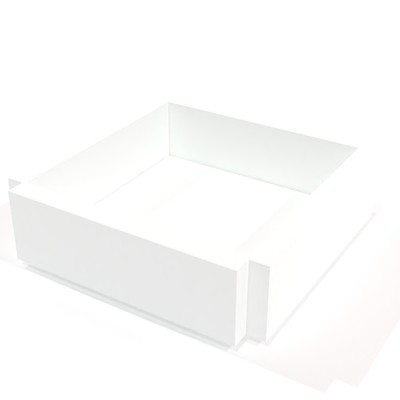}} \\[2.5pt]
    \qrow{\color{qours}\textbf{\method}\\\color{qours}(Ours)} & \qours{\includegraphics[width=\dimexpr\qw-1.2pt\relax,trim=0 32 0 32,clip]{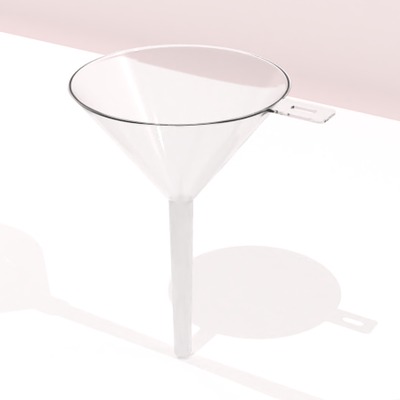}} & \qours{\includegraphics[width=\dimexpr\qw-1.2pt\relax,trim=0 32 0 32,clip]{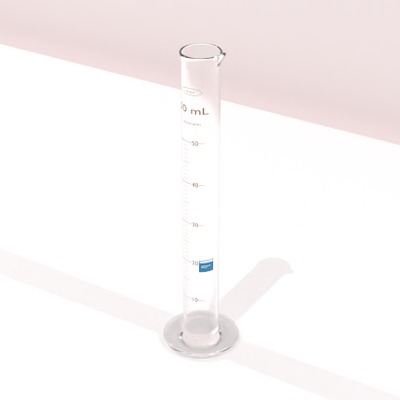}} & \qours{\includegraphics[width=\dimexpr\qw-1.2pt\relax,trim=0 32 0 32,clip]{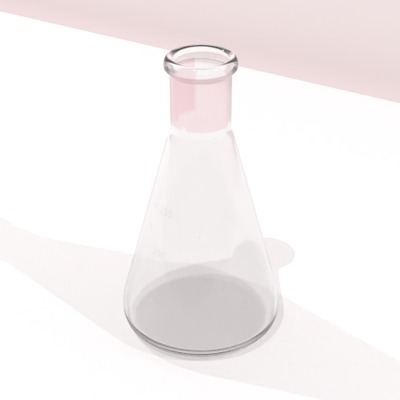}} & \qours{\includegraphics[width=\dimexpr\qw-1.2pt\relax,trim=0 32 0 32,clip]{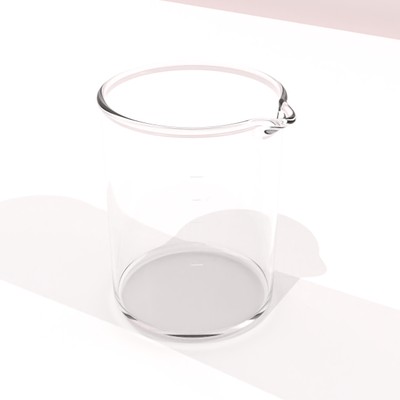}} & \qours{\includegraphics[width=\dimexpr\qw-1.2pt\relax,trim=0 32 0 32,clip]{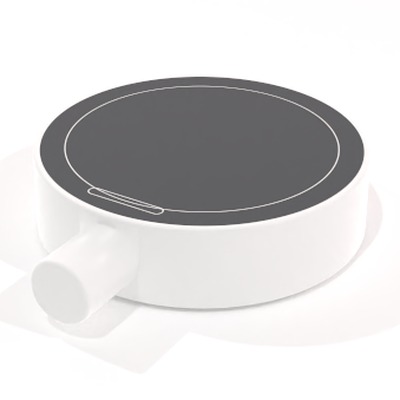}} & \qours{\includegraphics[width=\dimexpr\qw-1.2pt\relax,trim=0 32 0 32,clip]{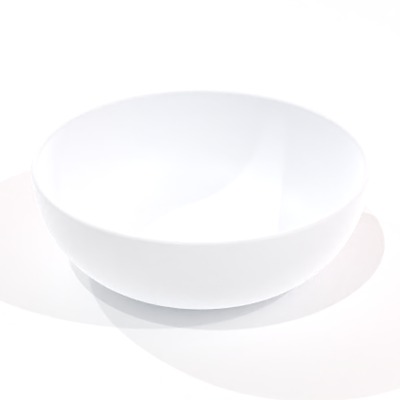}} & \qours{\includegraphics[width=\dimexpr\qw-1.2pt\relax,trim=0 32 0 32,clip]{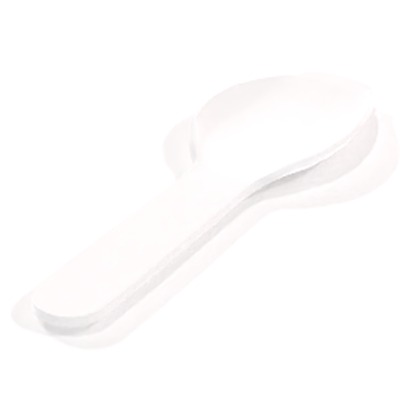}} & \qours{\includegraphics[width=\dimexpr\qw-1.2pt\relax,trim=0 32 0 32,clip]{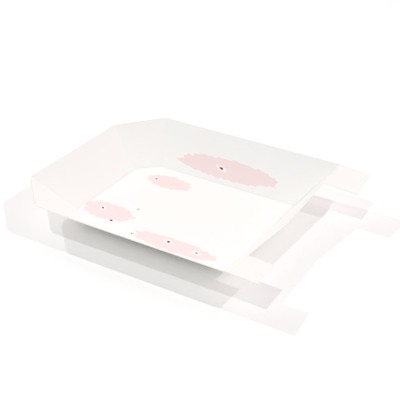}} \\
  \end{tabular}
  \caption{\textbf{Additional qualitative comparison.} Eight further instruments per scene that do not appear in Fig.~\ref{fig:qual}. For each scene, the first row shows the instrument cropped from a source video frame, and the remaining rows show the assets delivered by each method in the same RoboTwin tabletop scene, rendered with SAPIEN ray tracing (Real2Gym's MuJoCo models are rendered in MuJoCo). Gray cells mark instruments that a method did not recover.}
  \label{fig:qual_full}
\end{figure*}

\begin{figure*}[!t]
  \centering
  \setlength{\tabcolsep}{1pt}
  \renewcommand{\arraystretch}{0.8}
  \newcommand{\rw}{0.215\textwidth}
  \newcommand{\rrow}[1]{\rotatebox{90}{\parbox{0.16125\textwidth}{\centering\small #1}}}
  \newcommand{\rhead}[1]{\parbox[b]{\rw}{\centering\small #1}}
  \begin{tabular}{ccccc}
    & \multicolumn{3}{c}{Episode 1} & Episode 2 \\
    \cmidrule(lr){2-4}\cmidrule(lr){5-5}
    & \rhead{Open shield} & \rhead{Place paper} & \rhead{Press key} & \rhead{Pick up funnel} \\[1pt]
    \rrow{Real} & \includegraphics[width=\rw]{sec/figs/real2sim/open_shield_real.jpg} & \includegraphics[width=\rw]{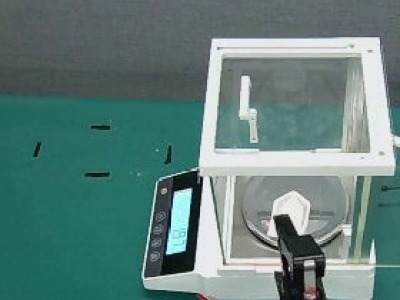} & \includegraphics[width=\rw]{sec/figs/real2sim/press_key_real.jpg} & \includegraphics[width=\rw]{sec/figs/real2sim/pick_funnel_real.jpg} \\
    \rrow{Sim} & \includegraphics[width=\rw]{sec/figs/real2sim/open_shield_sim.jpg} & \includegraphics[width=\rw]{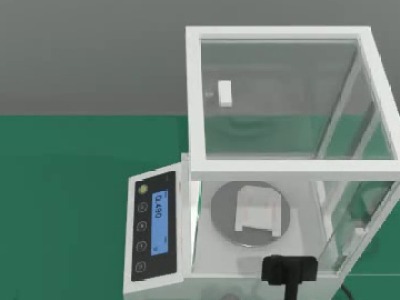} & \includegraphics[width=\rw]{sec/figs/real2sim/press_key_sim.jpg} & \includegraphics[width=\rw]{sec/figs/real2sim/pick_funnel_sim.jpg} \\[6pt]
    & Episode 2 & Episode 3 & Episode 4 & Episode 5 \\
    \cmidrule(lr){2-2}\cmidrule(lr){3-3}\cmidrule(lr){4-4}\cmidrule(lr){5-5}
    & \rhead{Insert funnel} & \rhead{Pour liquid} & \rhead{Pour powder} & \rhead{Scoop powder} \\[1pt]
    \rrow{Real} & \includegraphics[width=\rw]{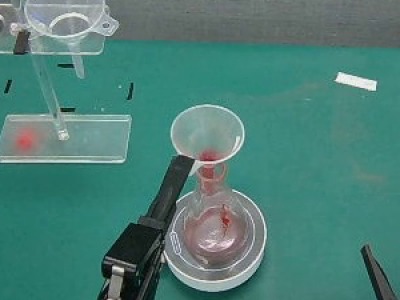} & \includegraphics[width=\rw]{sec/figs/real2sim/pour_liquid_real.jpg} & \includegraphics[width=\rw]{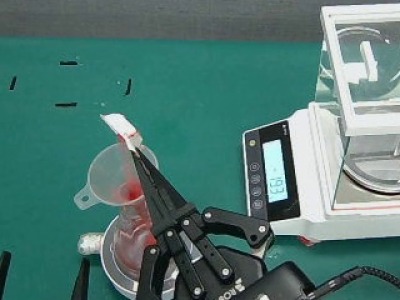} & \includegraphics[width=\rw]{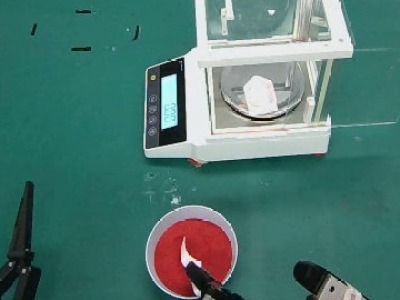} \\
    \rrow{Sim} & \includegraphics[width=\rw]{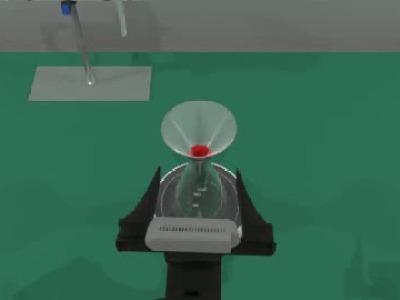} & \includegraphics[width=\rw]{sec/figs/real2sim/pour_liquid_sim.jpg} & \includegraphics[width=\rw]{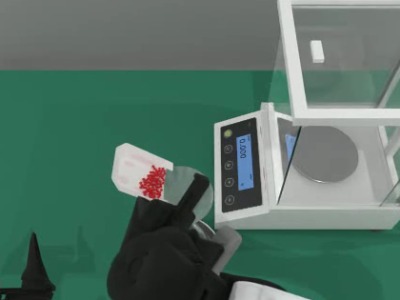} & \includegraphics[width=\rw]{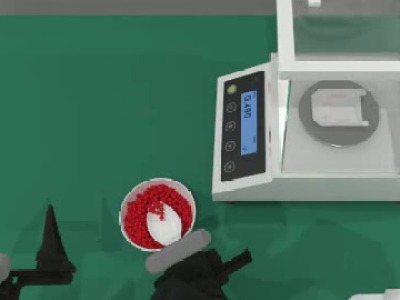} \\
  \end{tabular}
  \caption{\textbf{Real-to-sim reproduction in RoboTwin.} Five recorded robot episodes are reproduced in RoboTwin with assets generated by \method. Each column shows one atomic action: the real head-camera frame (top) and the simulated head-camera frame at the same action stage (bottom).}
  \label{fig:real2sim}
\end{figure*}
\section{Additional Results}
\label{app:results}

\noindent\textbf{Parts and articulation.}
On Scene~1, \method fully completes 28 required actions and partially completes six, yielding 31 action credits across the 41 required actions, compared with 18 for Codex + GPT-6-Astra, 6.5 for Codex + GPT-6-Sol, 2 for Real2Gym, and none for SimFoundry.
On Scene~2, \method completes all seven required actions.
Codex + GPT-6-Astra earns four action credits, while Codex + GPT-6-Sol and Real2Gym each earn three from two complete and two partial actions.
On Scene~1, the assets delivered by Codex + GPT-6-Astra lack the required movable components or release mechanisms for 23 of the 41 required actions, while those delivered by Codex + GPT-6-Sol lack them for 30.
These omissions include sliding draft-shield doors, spring-return buttons, and mechanisms for removing and replacing valve cores.
Real2Gym delivers 41 of the 54 required parts on Scene~1 and 22 of the 29 on Scene~2.
Its draft-shield doors are fixed to the balance body in both scenes, and the required cap, knob, and button motions are also absent on Scene~1.
Its two action credits on Scene~1 come from removing and reinserting the separatory-funnel stopper; its three credits on Scene~2 come from funnel removal and replacement operations involving the stand, graduated cylinder, and Erlenmeyer flask.
SimFoundry delivers 21 of the 54 required parts on Scene~1 and 18 of the 29 on Scene~2, and none of its deliveries completes a required action.

\noindent\textbf{Completeness and visual fidelity.}
On Scene~1, Codex + GPT-6-Astra matches all 18 reference instances but delivers two additional instances; Codex + GPT-6-Sol matches 16, Real2Gym matches 16 with three additional instances, and SimFoundry matches 11 with five additional instances.
On Scene~2, both Codex baselines match all 11 reference instances, Real2Gym misses the weighing paper, and SimFoundry matches nine with two additional instances.
The largest visual gains of \method over Codex + GPT-6-Astra on Scene~1 are in appearance and visible structure.
For appearance, \method scores 7.7, compared with 6.3 for Codex + GPT-6-Astra.
For visible structure, the corresponding ratings are 7.9 and 7.1.

\noindent\textbf{Ablations.}
The \emph{w/o feedback} variant loses 5.5 action credits on Scene~1 and one action and one part on Scene~2.
In its Scene~1 separatory funnel, liquid remains above the valve after opening, with no discharge through the lower outlet.
The \emph{images-input} variant matches 17 of the 18 reference instances on Scene~1, missing the stir bar, and delivers 48 of the 54 required parts with 20.5 action credits.
On Scene~2, its material sub-rating decreases from 8.3 to 7.6 and its appearance sub-rating from 7.7 to 7.2.
The two simulation-integration ablations have different completion profiles despite both earning 27 action credits on Scene~1: \emph{w/o adaptation} fully completes 19 actions and partially completes 16, whereas \emph{w/o joint presets} fully completes 16 and partially completes 22.
On Scene~2, \emph{w/o joint presets} fully completes the four funnel operations but only partially completes the three draft-shield door operations, yielding 5.5 action credits.

\noindent\textbf{Qualitative comparison.}
Fig.~\ref{fig:qual_full} shows eight additional instruments per scene that do not appear in Fig.~\ref{fig:qual}.
Columns follow the instances matched by the evaluator.

\noindent\textbf{Real-to-sim reproduction.}
Fig.~\ref{fig:real2sim} extends the representative operations in
Fig.~\ref{fig:tasks} to all five recorded robot episodes.
Using the same asset library produced by \method, GPT-6-Astra
reproduces the demonstrated operations in RoboTwin: opening the
draft shield, placing weighing paper and pressing a button in
Episode~1; picking up and inserting a funnel in Episode~2; and
pouring liquid, pouring powder, and scooping powder in Episodes~3
to~5, respectively.
The figure pairs real and simulated head-camera frames at
corresponding action stages.
Complete real-to-sim reproductions of all five episodes are
provided in the supplementary video.

\providecommand{\evalProtocolRef}{\ref{app:protocol}}
\providecommand{\evalPromptFinish}{}
\section{Evaluation Prompts}
\label{app:prompts}

This section presents the scoring instructions for the four
evaluation dimensions.
Within each scene, all methods are evaluated against the same
reference annotations using a common set of scoring criteria.
Judgments are based on source frames, standardized asset renders,
and simulation recordings, following the protocol in
Supp.~\ref{app:protocol}.

\onecolumn
\begingroup
\raggedbottom
\setlength{\tabcolsep}{3pt}
\newsavebox{\evalPromptTitle}
\newsavebox{\evalPromptBody}
\newlength{\evalPanelHeight}
\newlength{\evalBandBase}
\newcommand{\evalpromptheading}[1]{%
  \par\addvspace{3.4pt}\noindent\textbf{#1}\par\nobreak}

\newcommand{\evalRoundedFill}[3]{%
  \put(#3,0){\rule{\dimexpr#1-#3-#3\relax}{#2}}%
  \put(0,#3){\rule{#1}{\dimexpr#2-#3-#3\relax}}%
  \put(#3,#3){\circle*{\dimexpr#3+#3\relax}}%
  \put(\dimexpr#1-#3\relax,#3){\circle*{\dimexpr#3+#3\relax}}%
  \put(#3,\dimexpr#2-#3\relax){\circle*{\dimexpr#3+#3\relax}}%
  \put(\dimexpr#1-#3\relax,\dimexpr#2-#3\relax){\circle*{\dimexpr#3+#3\relax}}}

\makeatletter
\newenvironment{evalpromptpanel}[3]{%
  \par\addvspace{6pt}%
  \refstepcounter{subsection}\def\@currentlabelname{#1}\label{#2}%
  \addcontentsline{toc}{subsection}{\protect\numberline{\thesubsection}#1}%
  \sbox{\evalPromptTitle}{\normalfont\rmfamily\bfseries
    \fontsize{12}{14}\selectfont\strut\thesubsection.\enspace #1}%
  \begin{lrbox}{\evalPromptBody}%
    \begin{minipage}{\dimexpr\linewidth-22pt\relax}%
      \normalfont\renewcommand{\ttfamily}{\fontencoding{T1}\fontfamily{txtt}\selectfont}%
      #3\ttfamily\raggedright
      \setlength{\parindent}{0pt}\setlength{\parskip}{2pt}%
      \setlist[enumerate]{labelindent=0pt,align=left,labelwidth=1.4em,
        labelsep=.45em,leftmargin=!,itemindent=0pt,listparindent=0pt,
        topsep=2pt,itemsep=2pt,parsep=0pt}%
      \ignorespaces
}{%
      \par\end{minipage}%
  \end{lrbox}%
  \setlength{\evalBandBase}{\dimexpr\ht\evalPromptBody+\dp\evalPromptBody+16pt\relax}%
  \setlength{\evalPanelHeight}{\dimexpr\evalBandBase+\ht\evalPromptTitle+\dp\evalPromptTitle+10pt\relax}%
  \noindent\setlength{\unitlength}{1pt}%
  \begin{picture}(\linewidth,\evalPanelHeight)
    {\color{black!78}\evalRoundedFill{\linewidth}{\evalPanelHeight}{5pt}}%
    \put(.5pt,.5pt){{\color{black!14}%
      \evalRoundedFill{\dimexpr\linewidth-1pt\relax}{\dimexpr\evalPanelHeight-1pt\relax}{4.5pt}}}%
    \put(.5pt,.5pt){{\color{black!3}%
      \evalRoundedFill{\dimexpr\linewidth-1pt\relax}{\dimexpr\evalBandBase-.5pt\relax}{4.5pt}}}%
    \put(.5pt,\dimexpr\evalBandBase-5pt\relax){\color{black!3}\rule{\dimexpr\linewidth-1pt\relax}{5pt}}%
    \put(.5pt,\evalBandBase){\color{black!55}\rule{\dimexpr\linewidth-1pt\relax}{.35pt}}%
    \put(11pt,\dimexpr\evalBandBase+5pt+\dp\evalPromptTitle\relax){\usebox{\evalPromptTitle}}%
    \put(11pt,\dimexpr8pt+\dp\evalPromptBody\relax){\usebox{\evalPromptBody}}%
  \end{picture}\par
}
\makeatother

\begin{evalpromptpanel}{Prompt A --- Instance Completeness}{app:prompt_a}{\fontsize{8.8}{10.2}}
\upshape
Evaluate how completely the delivered assets represent the instruments and utensils used in the source laboratory video. Score the scene from 0 to 100 using instance-level intersection over union (IoU).

\textbf{Inputs:} \texttt{source\_video} (one video or source-manifest clips), \texttt{asset\_inventory} (delivery rows and asset paths), and an asset viewer; a shared \texttt{video\_reference.json} and a delivery relationship document may also be supplied.

\evalpromptheading{1. Establish the video reference}

Video annotations are an external input. Use the saved \texttt{video\_reference.json}, preserving instance IDs and visible features. Inspect its source evidence before opening generated assets. Do not rebuild its inventory based on this delivery or read prior evaluations\textquotesingle{} matches, judgments or scores.

Only when no shared reference is supplied, watch the source video before opening the asset inventory and establish one. Assign one reference ID to each distinct task-relevant instrument or utensil. Record its category, distinguishing cues, and a clip ID with a clip-local timestamp or time range. Preserve the source manifest\textquotesingle s clip order; filename order alone does not establish a continuous chronology. Track the same object across frames. Count two separately used beakers as two instances; record a bottle cap or a push button as a part of its instrument. Use equipment\_only scope: include laboratory instruments, utensils, stands and clamps; exclude people, robots, benches and background.

Save the reference as \texttt{video\_reference.json} and reuse it for every method evaluated on this video. For use by Prompts B\textendash{}D, include visible appearance and material cues, required parts and motions, and any separatory funnel used for valve-controlled drainage.

\evalpromptheading{2. Match the generated instances}

Open the delivered assets and enumerate physical instrument instances. A package may contain two separate instruments; several export files may represent one instrument. Match video instances to generated instances one to one. Use distinguishing features when available. For visually indistinguishable instances of the same type, pair as many as are present on both sides.

The VLM must inspect actual renders and component views to check whether different inventory rows are complementary parts of one instrument. Use the supplied relationship document as supporting delivery information and map its identifiers to this inventory; the inventory still selects the actual asset paths and versions. A separately delivered bottle body and its cap may form one instance. Retain the shared reference\textquotesingle s boundaries for independent instruments such as a funnel and stand or a hotplate and loose stir bar. Do not deduplicate genuine repeated instruments or parts solely by name or category, and never use one physical part to satisfy two reference instances.

Account for every row and every physical component. A row count is not the generated instance count \texttt{G}. Grouping changes evaluation identity only; do not edit the source inventory, geometry, materials or mechanisms, and do not invent connections. Resolve ambiguous ownership with targeted evidence and retain any remaining uncertainty.

Match by instrument identity. For example, a recognizable beaker with the wrong surface finish still matches the reference beaker; its finish is scored in Prompt B.

\evalpromptheading{3. Score and report}

Let \texttt{V} be the reference count, \texttt{G} the generated count, and \texttt{M} the number of matched pairs.

\par\noindent
\begin{tabular}{@{}p{.32\linewidth}p{\dimexpr.68\linewidth-2\tabcolsep\relax}@{}}
  \toprule
  \textbf{Quantity} & \textbf{Calculation} \\
  \midrule
  Matched (\texttt{TP}) & \texttt{M} \\
  Missing (\texttt{FN}) & \texttt{V\ \ensuremath{-}\ M} \\
  Extra (\texttt{FP}) & \texttt{G\ \ensuremath{-}\ M} \\
  Union (\texttt{U}) & \texttt{V\ +\ G\ \ensuremath{-}\ M} \\
  Scene score & \texttt{A\ =\ 100\ \ensuremath{\times}\ M\ /\ U} \\
  \bottomrule
\end{tabular}
\par

If only one inventory is empty, score 0. If both are empty, report \texttt{N/A}. Example: 10 reference instances, 10 generated instances, and 9 matches give \texttt{100\ \ensuremath{\times}\ 9\ /\ 11\ =\ 81.82}.

For ambiguous-looking pairs, identify the visual cue or shared category count used for the match.
\end{evalpromptpanel}

\clearpage

\begin{evalpromptpanel}{Prompt B --- Visual Fidelity}{app:prompt_b}{\fontsize{8.0}{8.7}}
\upshape
Compare the delivered laboratory assets with the instruments visible in the source video. Report a scene score out of 100: \textbf{appearance 35, material appearance 35, visible structure 30}. Ratings are \textbf{integers 1\textendash{}10, with 0 for missing instances or wholly inconsistent defining features}.

\textbf{Inputs:} \texttt{source\_video}, \texttt{asset\_inventory} (instance IDs and asset paths), and an asset viewer or renderer. Reuse the supplied shared \texttt{video\_reference.json} unchanged, including IDs, visible features and applicability. Use this evaluation\textquotesingle s instance matching from Prompt A, and inspect all grouped members of each instance. A supplied delivery relationship document helps identify complementary components across inventory rows but does not establish visual correctness. Never import prior methods\textquotesingle{} matches, ratings or comparison boards.

\evalpromptheading{Inspect and rate}

Choose frames showing each reference instrument clearly. Rotate or zoom the generated asset to inspect the corresponding surfaces and parts. Judge the visible instrument in comparable views. Give \textbf{one 1\textendash{}10 integer rating per applicable subscore per reference instance}, retaining \textbf{0} for the cases defined below. A physical instrument assembled from multiple delivery entries receives one set of ratings covering all its actual members; do not average separate row-level scores or count a shared part twice. Inspect actual source images and generated renders, not metadata alone. Use the protocol\textquotesingle s gray background, original materials, four views and necessary component details; do not repair or restyle the asset.

\par\noindent
\begin{tabular}{@{}>{\raggedright\arraybackslash}p{.18\linewidth}>{\centering\arraybackslash}p{.09\linewidth}>{\raggedright\arraybackslash}p{\dimexpr.73\linewidth-4\tabcolsep\relax}@{}}
  \toprule
  \textbf{Subscore} & \textbf{Weight} & \textbf{What to judge} \\
  \midrule
  Appearance & 35 & Overall silhouette, visible proportions, body shape, surface graphics and markings. Compare printed patterns, colored bands, labels, panel icons, and printed graduations by presence, design, color, placement, and coverage. Describe both shape and markings in the evidence note. \\
  Material appearance & 35 & Base-material color, transparency or translucency, opacity, gloss, and frosting. Compare the regions carrying each finish. A bottle\textquotesingle s amber tint belongs here; an amber band printed on the bottle belongs under appearance. \\
  Visible structure & 30 & Three-dimensional form and placement of features: neck and opening, stopcock and outlet, raised thread profile, and panel\textendash{}knob layout. Score the printed arrow around a knob under appearance; score the knob\textquotesingle s shape and placement here. \\
  \bottomrule
\end{tabular}
\par

\par\noindent
\begin{tabular}{@{}>{\raggedright\arraybackslash}p{.085\linewidth}>{\raggedright\arraybackslash}p{\dimexpr\linewidth-.085\linewidth-2\tabcolsep\relax}@{}}
  \toprule
  \textbf{Rating} & \textbf{Anchor} \\
  \midrule
  10 & Main features and all clearly distinguishable details agree; no definite discrepancy is visible at the supported evidence resolution. \\
  9 & Main features and almost all distinguishable details agree, with very few slight discrepancies. \\
  8 & Main features agree, with small discrepancies in a few local details. \\
  7 & Main features agree, with clear discrepancies in several local details; overall correspondence remains strong. \\
  6 & Most main features agree, with a clear discrepancy in at least one main feature or substantial discrepancies in several local details. \\
  5 & Some main features agree, alongside several conspicuous differences; fidelity is intermediate. \\
  4 & Only some main features and a few details agree; conspicuous differences dominate. \\
  3 & The category and a few distinguishing features agree, but most main features differ. \\
  2 & Correspondence is largely limited to the category, with few features specific to the reference instance. \\
  1 & Only minimal generic resemblance remains; almost none of the reference-specific features are reproduced. \\
  0 & The required instance is missing, or this subscore\textquotesingle s defining visible features are wholly inconsistent. \\
  \bottomrule
\end{tabular}
\par

For example, a correctly shaped bottle missing a clearly visible wide band has an appearance discrepancy: identify whether that band is a main distinguishing feature or a local detail and apply the corresponding anchor. An invented pattern on a clearly plain surface is also an appearance difference. When label text is unreadable, compare the visible label shape, color, and placement. Rate graphics by their rendered appearance, regardless of how they were authored. Liquid contents are not bottle-wall tint. Record the agreements and discrepancies supporting each subscore. Rate directly on these anchors; do not invent decimal precision.

\evalpromptheading{Aggregate and report}

Reuse applicability from the shared source reference. Mark a subscore \texttt{N/A} only when its reference features cannot be seen, not because the generated asset or its review is missing. Score a missing generated instance \textbf{0 on each reference-supported subscore}, retaining it in the denominator.

For each subscore, average its applicable reference-instance ratings. With means \texttt{a}, \texttt{m}, and \texttt{s}, compute \textbf{\texttt{B\ =\ 35\ \ensuremath{\times}\ a/10\ +\ 35\ \ensuremath{\times}\ m/10\ +\ 30\ \ensuremath{\times}\ s/10}}. If an entire subscore lacks reference evidence, retain the relative weights of the applicable subscores and rescale to 100: \texttt{B\ =\ 100\ /\ sum(applicable\ weights)\ \ensuremath{\times}\ sum(weight\ \ensuremath{\times}\ mean/10)}. Report that coverage. If none applies, report \texttt{N/A}. Use unrounded means.

Save each completed instance review with reference and generated IDs, the three ratings, observations and source clip-local timestamps/render IDs.
\end{evalpromptpanel}

\clearpage

\begin{evalpromptpanel}{Prompt C --- Parts and Articulation}{app:prompt_c}{\fontsize{8.8}{10.2}}
\upshape

Evaluate the delivered laboratory assets in the simulator. Award \textbf{50 points for required parts and 50 points for articulation}.

\textbf{Inputs:} \texttt{source\_video}, external video annotations, \texttt{asset\_inventory}, and \texttt{sim\_env}. Use the \texttt{video\_reference.json} IDs and required parts/actions; use this evaluation\textquotesingle s matching and grouping. External annotations may include video observations and explicitly identified common-knowledge requirements even without filmed movement. Do not invent timestamps, mechanisms, strokes or tolerances. Apply user exclusions before determining required actions: excluded operations carry no credit and no denominator entry; static part scope is independent. No equipment-specific requirements are built into the evaluator. Reuse the same annotation revision across methods, never their judgments or trials.

Inspect every actual member of an instance whose components span inventory rows. Use the supplied relationship document and visual evidence to establish membership, retaining each member\textquotesingle s original asset path, part IDs and mechanisms. Count each required part and action once per reference instance. A relationship description is not proof that a part or joint was delivered. Grouping does not repair an asset: use only declared, physically supported assembly relationships and protocol-permitted initial placement/support; do not invent connections, constraints, masses or missing geometry.

\evalpromptheading{1. Required parts \textemdash{} 50 points}

Inspect each asset from views that reveal its required parts. Score each distinct required part once: \textbf{1} for present, \textbf{0} for absent. Inspect the delivered geometry or rendering. A recognizable but poorly finished part earns its presence credit. A missing reference instrument contributes 0 for each of its required parts.

\texttt{part\_points\ =\ 50\ \ensuremath{\times}\ present\ required\ parts\ /\ total\ required\ parts}

\evalpromptheading{2. Articulation \textemdash{} 50 points}

For each required part\textendash{}action pair, actuate the delivered part in the simulator and observe its motion relative to the instrument body. For example, a knob turns about its shaft, a button moves inward relative to the panel, and a sliding door moves along its guide. Reuse the same action and viewing conditions across methods.

Use external \texttt{completion\_criteria} to assess complete operations: press/release with source spring return, open/close to the defined endpoint, loosen/adjust/relock, or detach/reseat. A \texttt{phases} sequence still maps to one \texttt{reference\_id/action\_id} and one action credit. For a cross-instrument \texttt{interface}, record both participants and count the operation once under its declared owner; observe movement relative to the mating instrument. Whole-instrument transport alone earns no credit. Actual source-declared screw/clamp release is allowed; arbitrary constraint deletion or invented assembly is not.

Use the saved native protocol and this delivery\textquotesingle s actual joint coordinates, mechanisms and finite-force drives. Observe the continuous recording; a render export, joint declaration or command exit code is not evidence of successful relative motion. Preserve source geometry, materials, colliders, mass and inertia. Grouping complementary files does not by itself earn action credit.

If incomplete travel may be caused by the test drive, request diagnostic trials: the evaluator creates separate trials using the saved finite list of gain profiles, retaining force limits, friction, geometry and baseline evidence. Review controller behavior and every recording before attributing failure to the asset; record the selected trial and reason in the action judgment. Do not auto-select the highest score or change all joints to one stronger gain.

\par\noindent
\begin{tabular}{@{}>{\raggedright\arraybackslash}p{.15\linewidth}>{\raggedright\arraybackslash}p{\dimexpr\linewidth-.15\linewidth-2\tabcolsep\relax}@{}}
  \toprule
  \textbf{Action credit} & \textbf{Observed outcome} \\
  \midrule
  1 & The part moves in the intended way and completes the operation. \\
  0.5 & The motion type is correct, but travel is visibly insufficient, the part sticks, or the operation completes only partially. \\
  0 & The required part is missing or stationary, the motion type is wrong, or the whole instrument moves without the intended relative part motion. \\
  \bottomrule
\end{tabular}
\par

\texttt{motion\_points\ =\ 50\ \ensuremath{\times}\ sum\ of\ action\ credits\ /\ number\ of\ required\ part\textendash{}action\ pairs}

Count required actions even when their part or instrument is missing. Instruments with no expected moving parts contribute only to the parts assessment. If the entire scene has no required actions, report articulation as \texttt{N/A} and compute \texttt{C\ =\ 100\ \ensuremath{\times}\ present\ required\ parts\ /\ total\ required\ parts}, labeled \textbf{parts only}. Otherwise, \texttt{C\ =\ part\_points\ +\ motion\_points}.
\end{evalpromptpanel}

\clearpage

\begin{evalpromptpanel}{Prompt D --- Valve-Controlled Drainage}{app:prompt_d}{\fontsize{8.8}{10.2}}
\upshape

Test whether each separatory funnel required by the source video can retain liquid with its stopcock closed and discharge liquid through its lower outlet when the stopcock opens. Score each applicable instance \textbf{100 or 0}.

\textbf{Inputs:} \texttt{source\_video}, \texttt{asset\_inventory} (instance IDs and asset paths), and \texttt{sim\_env} with an \textbf{MPM liquid solver}. Reuse supplied \texttt{video\_reference.json} funnel requirements unchanged; use this evaluation\textquotesingle s instance matching and grouping from Prompt A. Do not import another method\textquotesingle s trial, calibration or score. If the reference contains no separatory funnel, report \textbf{D = N/A}.

A funnel body, stopper or stopcock may be separate inventory entries. Confirm their ownership using actual renders, delivery records and the supplied relationship document, and retain all members\textquotesingle{} real paths in the grouping map. Test the resulting physical instance once, not once per file. Use only mechanisms actually delivered; membership does not supply a missing joint, seal or part. Do not edit the source inventory, geometry, materials, colliders, mass, inertia or mechanisms, and do not substitute assets from the relationship document\textquotesingle s directory.

\evalpromptheading{Run the trial}

\begin{enumerate}[label=\arabic*.]
  \item Load the delivered funnel in its normal supported working position using the saved native protocol. The declared action fixture may support the glass and stand roots while leaving the valve movable. Start with liquid in the funnel and the stopcock closed.
  \item Frame the stopcock and lower outlet together. Record one continuous clip of the closed and open phases.
  \item Observe the outlet during the closed phase. Then actuate the stopcock to open it and observe where liquid exits. Drive the stopcock; let the MPM solver compute the liquid motion through the delivered geometry.
  \item Use the same liquid settings, actuation protocol, and observation windows for every method.
\end{enumerate}

Use RoboTwin\textquotesingle s native scene and CUDA MPM, 1/240-second rigid steps, 50\% actual-cavity fill, 3 mm spacing, 32 MPM substeps, and continuous 2-second closed / 1-second opening / 4-second open phases. Calibrate horizontal-closed and vertical-open lever states from this delivery\textquotesingle s coordinates; do not copy another asset\textquotesingle s q values. Retain native mesh contact and swept CCD; do not add an ideal seal, prescribed emitter or geometry repair. Inspect actual continuous native footage or the permitted saved-state surface replay. Particle counts and tool success are supporting data, not substitutes for visible flow.

\evalpromptheading{Score the observed behavior}

\par\noindent
\begin{tabular}{@{}>{\raggedright\arraybackslash}p{.18\linewidth}>{\raggedright\arraybackslash}p{\dimexpr\linewidth-.18\linewidth-2\tabcolsep\relax}@{}}
  \toprule
  \textbf{Instance score} & \textbf{Criterion} \\
  \midrule
  100 & No sustained discharge is visible from the lower outlet while closed; after opening, liquid visibly flows through that outlet. \\
  0 & The required funnel or stopcock is missing, the outlet drains continuously while closed, opening produces no outlet flow, or liquid only escapes through an unintended path such as the vessel wall. \\
  \bottomrule
\end{tabular}
\par

A stopcock that rotates but leaves the outlet blocked earns \textbf{0} for drainage. Rotation is a separate articulation result in Prompt C. Record the timestamps of both phases and a short observation for each.

Compute \texttt{D\ =\ 100\ \ensuremath{\times}\ passing\ funnels\ /\ applicable\ funnels}.
\end{evalpromptpanel}

\endgroup
\twocolumn
\evalPromptFinish

\end{document}